\documentclass[final, onefignum,onetabnum]{siamart190516}
\usepackage{lipsum}
\usepackage{amsfonts}
\usepackage{graphicx}
\usepackage{epstopdf}
\ifpdf
  \DeclareGraphicsExtensions{.eps,.pdf,.png,.jpg}
\else
  \DeclareGraphicsExtensions{.eps}
\fi

\newsiamremark{remark}{Remark}
\newsiamremark{hypothesis}{Hypothesis}
\crefname{hypothesis}{Hypothesis}{Hypotheses}
\newsiamthm{claim}{Claim}

\headers{Globally Convergent Multilevel Training Of ResNets}{A. Kopani\v{c}\'akov\'a, R. Krause}

\title{Globally Convergent Multilevel Training of Deep Residual Networks\thanks{Submitted to the editors DATE.
\funding{This work was funded by the Swiss National Science Foundation (SNF) under the project ML$^2$ (grant no.~197041) and by Platform for Advanced Scientific Computing (PASC) under the project EXATRAIN.}}}

\author{Alena Kopani\v{c}\'akov\'a \thanks{Euler Institute, Universit{\`a} della Svizzera italiana
  (\email{alena.kopanicakova@usi.ch}, \url{http://usi.to/r8u}).}
\and Rolf Krause\footnotemark[2]}

\usepackage{amsopn}

 \usepackage[T1]{fontenc}
\usepackage[latin1]{inputenc}
\usepackage{caption}
\usepackage{algorithm,algpseudocode}
\usepackage{graphicx,url}
\usepackage{amsmath,amssymb,amsopn}
\usepackage{tabularx}
\usepackage{color}
\usepackage{calc}
\usepackage{caption}
\usepackage{longtable}
\usepackage{multirow}
\usepackage{mathrsfs}
\usepackage{array}
\usepackage{hyperref}
\usepackage{tikz}
\usepackage{booktabs}
\usepackage{pgfplotstable}
\usepackage{pgfplots}
\usepackage[outline]{contour}
\usepackage[colorinlistoftodos,prependcaption]{todonotes}
\usepackage{calrsfs}
\usepackage{graphicx}
\usepgfplotslibrary{groupplots}
\usetikzlibrary{matrix}
\usetikzlibrary{automata, external, shapes.geometric, fit}
\pgfplotsset{compat=1.13}

\def \bv{\vec{b}}
\def \cv{\vec{c}}

\def \gv{\vec{g}}

\def \nv{\vec{n}}

\def \sv{\vec{s}}

\def \vv{\vec{v}}

\def \xv{\vec{x}}
\def \yv{\vec{y}}
\def \zv{\vec{z}}
\def \thetav{\vec{\theta}}

\def \rm{\mat{r}}

\def \Bm{\mat{B}}

\def \Dm{\mat{D}}

\def \Im{\mat{I}}

\def \Qm{\mat{Q}}
\def \Pm{\mat{P}}
\def \Rm{\mat{R}}

\def \Wm{\mat{W}}

\def \R{\mathbb{R}}			\def \N {\mathbb{N}}

\def \D{\pazocal{D}}			\def \H{\pazocal{H}}			\def \L{\pazocal{L}}			 		\def \D{\pazocal{D}}

\DeclareMathAlphabet{\pazocal}{OMS}{zplm}{m}{n}

\renewcommand{\vec}[1]{\boldsymbol{#1}}
\newcommand{\mat}[1]{\boldsymbol{\mathrm{#1}}}

\algnewcommand\algorithmiconput{\textbf{Constants:}}
\algnewcommand\algorithmicinput{\textbf{Input:}}
\algnewcommand\algorithmicoutput{\textbf{Output:}}
\algnewcommand{\algorithmicgoto}{\textbf{go to}}

\algnewcommand\Constants{\item[\algorithmiconput]}
\algnewcommand\Input{\item[\algorithmicinput]}\algnewcommand\Output{\item[\algorithmicoutput]}\algnewcommand{\Goto}[1]{\algorithmicgoto~\ref{#1}}

\definecolor{myblack}{RGB}{53, 53, 53}
\definecolor{myblue}{RGB}{40, 75, 99}
\definecolor{myred}{RGB}{192, 50, 33}
\definecolor{myyellow}{RGB}{255, 166, 48}
\definecolor{mywhite}{RGB}{240, 237, 238}
\definecolor{mygreen}{RGB}{0, 102, 0}

\definecolor{green1}{RGB}{9, 82, 86}
\definecolor{green2}{RGB}{8, 127, 140}
\definecolor{green3}{RGB}{6, 167, 125}
\definecolor{green4}{RGB}{79, 109, 122}
\definecolor{green5}{RGB}{192, 214, 223}
\definecolor{violet}{RGB}{26,69,131}

\definecolor{checkgreen}{rgb}{0,0.6,0}
\definecolor{phase1}{rgb}{0.008,0.655,1.000}
\definecolor{phase2}{rgb}{0.016,0.75,0.700}
\definecolor{phase3}{rgb}{0.929,0.35,0.700}
\definecolor{icsyellow}{cmyk}{0.00,0.11,0.53,0.00}

\definecolor{blackmy}{RGB}{38, 70, 83}

\definecolor{bluemy}{RGB}{39, 125, 161}
\definecolor{greenmy}{RGB}{42, 167, 143}
\definecolor{yellowmy}{RGB}{233, 196, 106}
\definecolor{brownmy}{RGB}{244, 162, 97}
\definecolor{redmy}{RGB}{249, 65, 68}

\usetikzlibrary{patterns}
\usepackage[normalem]{ulem}
\usepackage{soul,xcolor}
\usepackage[many]{tcolorbox}

\usepackage{pgfplots}
\usetikzlibrary{intersections}
\usepgfplotslibrary{fillbetween}
\usepackage{bbm}

\ifpdf
\hypersetup{
  pdftitle={Globally Convergent Multilevel Training Of ResNets},
  pdfauthor={A. Kopani\v{c}\'akov\'a, and R. Krause}
}
\fi

\definecolor{darkbluemy}{RGB}{65, 59, 147}
\definecolor{lightbluemy}{RGB}{71, 139, 194}
\definecolor{greenmy}{RGB}{98, 173, 153}
\definecolor{darkorangemy}{RGB}{230, 142, 52}
\definecolor{lightorangemy}{RGB}{217, 172, 59}

\begin{document}

\maketitle

\begin{abstract}
We propose a globally convergent multilevel training method for deep residual networks (ResNets). 
The devised method can be seen as a novel variant of the recursive multilevel trust-region (RMTR) method, which operates in hybrid (stochastic-deterministic) settings by adaptively adjusting mini-batch sizes during the training. 
The multilevel hierarchy and the transfer operators are constructed by exploiting a dynamical system's viewpoint, which interprets forward propagation through the ResNet as a forward Euler discretization of an initial value problem. 
In contrast to traditional training approaches, our novel RMTR method also incorporates curvature information on all levels of the multilevel hierarchy by means of the limited-memory SR1 method.
The overall performance and the convergence properties of our  multilevel training method are numerically investigated using examples from the field of classification and regression.
\end{abstract}



\begin{sloppypar}
\section{Introduction}
Deep residual networks (ResNets)~\cite{he2016deep, he2016identity} are widely used network architectures, as they demonstrate state-of-the-art performance in complex statistical learning tasks.
The ResNet architecture utilizes a so-called shortcut connection, which allows for the propagation of a signal directly from one block to another. 
The use of this shortcut connection enabled the training of networks with hundreds or even thousands of layers, which in turn provided an increase in network approximation power~\cite{haastad1991power}. 
Indeed, since the inception of ResNets, the performance of many learning tasks, e.g., from the field of computer vision~\cite{jung2017resnet, chen2017deeplab}, has been significantly improved.

Despite their remarkable performance, ResNets suffer from a long training time.
This is due to the fact that the convergence properties of many optimizers tend to deteriorate with the increasing network depth. 
Additionally, the cost associated with the forward-backward propagation (gradient evaluation) increases linearly with respect to the number of layers~\cite{chauvin1995backpropagation}.
To mitigate the difficulty, different strategies have been proposed, e.g.,~networks with stochastic depth~\cite{huang2016deep}, spatially adaptive architectures~\cite{figurnov2017spatially}, or mollifying networks~\cite{gulcehre2016mollifying}.
In this work, we propose to accelerate the training of ResNets by introducing a novel multilevel training strategy. 
The proposed method can be seen as an extension of the multilevel trust-region method~\cite{Gratton2008recursive, Gross2009}.
The design of the proposed training method is motivated by the observations discussed in the following paragraphs.

The training of ResNets is typically performed using variants of the stochastic gradient (SGD) method~\cite{robbins1951stochastic}, which construct search directions using a stochastic gradient estimator. 
Although these methods have a low computational cost per iteration, their convergence properties rely heavily on the choice of hyper-parameters. 
More precisely, it is important to carefully select a sequence of diminishing step-sizes to ensure convergence to a solution. 
To reduce the dependency of the solution method on the hyper-parameters, we propose to employ a trust-region based optimizer.
The sequence of step-sizes is then determined automatically by the trust-region method~\cite{Conn2000trust}.

Trust-region methods have been originally developed for solving deterministic optimization problems. 
In particular, they are of interest for non-convex optimization problems, such as ones considered in this work, as they offer global convergence guarantees. 
More recently, there has been  growing interest in developing stochastic trust-region methods. 
The pursued strategies can be roughly classified into three groups, depending on the way the sampling is performed to obtain approximate information about the objective function and its derivatives.  
The first two groups consist of  methods, which evaluate the objective function exactly, 
but employ sub-sampled gradient and Hessian information~\cite{erway2020trust, gratton2018complexity}, or use exact gradient and sub-sample only curvature information~\cite{xu2020newton, xu2020second}.

In contrast, the methods from the third group employ only stochastic estimates of the objective function and its derivatives~\cite{bellavia2018levenberg, blanchet2019convergence, chen2018stochastic}. 
This gives rise to computationally efficient numerical methods of stochastic nature. 
However, to preserve the global convergence properties of the trust-region method, the objective function  and gradient have to be estimated with increasing accuracy. 
For finite sum problems, the accuracy of the estimates can be increased by enlarging the sample sizes~\cite{bollapragada2018adaptive}. 
In this work, we follow the approach proposed in~\cite{mohr2019adaptive} and utilize a dynamic sample size (DSS) strategy, which adaptively increases the sample sizes during the training process. 
Thus, we obtain a hybrid (stochastic-deterministic) method, which takes advantage of small-batches at the beginning of the training process. 
As training progresses, the mini-batch size is adaptively increased, which ensures convergence to a solution.

Unfortunately, the convergence rate of the iterative methods, such as trust-region, often deteriorates with the network depth, i.e.,~the number of iterations required to reach the desired tolerance grows rapidly with the number of parameters.
Multilevel methods are known to be optimal solvers for many problems, in the sense that their convergence rate is often independent of the problem size, and that the number of required arithmetic operations grows proportionally with the number of unknowns. 
These methods have originally been developed for numerically solving linear elliptic partial differential equations (PDEs)~\cite{briggs2000multigrid}.
Full approximation scheme (FAS)~\cite{brandt1977multi} and  nonlinear multigrid (NMG)~\cite{hackbusch2013multi} 
have been proposed to extend the multigrid methods to  nonlinear PDEs. 
In the last decades, several nonlinear multilevel minimization techniques have emerged, e.g.,~the multilevel line-search method (MG/OPT)~\cite{nash2014properties}, the recursive multilevel trust-region method (RMTR)~\cite{Gratton2008recursive, gratton2008_inf}, monotone multigrid method~\cite{RKornhuber_1997b, RKornhuber_RKrause_2001} or higher-order multilevel optimization strategies (MARq)~\cite{calandra2021high, calandra2019approximation}. 
In this work, we utilize the RMTR method, which is designed for solving non-convex optimization problems.
By now, several variants of the RMTR method have been proposed and investigated in the literature~\cite{kragel2005streamline, ziems2011adaptive, ulbrich2017adaptive,  kopanivcakova2020recursive, subdivision_kopanicakova, chegini2021efficient, kopanivcakova2021multilevel, zulian2020large}, but, to the best of our knowledge, the method has not been extended into stochastic settings nor it has been applied for training of deep neural networks.

The implementation of the RMTR method requires two main components: a multilevel hierarchy and transfer operators. 
In this work, we construct both by leveraging the dynamical system's viewpoint~\cite{haber2018learning, weinan2017proposal}, which interprets a forward propagation through the ResNet as the discretization of an initial value problem.
The training process can then be formulated as the minimization of a time-dependent optimal control problem.
As a consequence, a hierarchy of ResNets with different depths can be obtained by discretizing the same optimal control problem with different discretization parameters (time-steps).
The RMTR method can then accelerate the training of the deepest ResNet by internally training the shallower networks.

Several authors have recently pursued the development of multilevel training methods for ResNets. 
For example, Haber et al.~proposed two multilevel training approaches in~\cite{haber2018learning}. 
In the first approach, the multilevel hierarchy was created by changing an image resolution, while the second approach utilized the dynamical system's viewpoint. 
Both methods employed the cascadic multigrid approach and utilized the multilevel hierarchy of ResNets only to gradually initialize the network parameters, see also~\cite{chang2017multi, cyr2019multilevel} for additional numerical study.  
Furthermore, Wu et al.~\cite{wu2020multigrid} proposed a multilevel training for video sequences. 
The multilevel methods were also explored in the context of layer-parallel training in~\cite{gunther2018layer, kirby2020layer}. 
Let us note eventually that a variant of the multilevel line-search method was presented in~\cite{Kopanicakova_2020c}.
Similar to the proposed RMTR method, the method utilized the dynamical system's viewpoint in order to construct a multilevel hierarchy and transfer operators. 
In contrast to our RMTR method, its performance relied on a large number of hyper-parameters. 
More precisely, a learning rate and its decay factor had to be selected carefully on each level of the multilevel hierarchy in order to ensure convergence. 
Moreover, none of the aforementioned training methods incorporated curvature information nor provided global convergence guarantees.

This paper is organized as follows:  
\Cref{sec:resnet} provides a brief introduction to supervised learning, with a particular focus on the continuous optimal control training framework. 
In \cref{sec:rmtr_resnet}, we describe the RMTR method and discuss how to obtain a multilevel hierarchy and transfer operators in the context of ResNets. 
\Cref{sec:dynamic_sampling} proposes an extension of the RMTR method into hybrid (stochastic-deterministic) settings, which is achieved using a dynamic sample size strategy. 
\Cref{sec:numerical_experiments} describes various numerical examples, which we employ for testing the proposed multilevel training method.
Finally, \cref{sec:num_results} demonstrates the overall performance of the proposed training method. 
In the end, the summary and possible future work are discussed in \cref{sec:conclusion}.

 \section{Supervised learning as a continuous optimal control problem}
\label{sec:resnet}
In this section, we provide a brief introduction to supervised learning.  
To this aim, we consider a dataset $\D = \{ (\xv_s, \cv_s) \}_{s=1}^{n_s}$, which contains $n_s$ samples. 
Each sample is defined by input features $\xv_s \in \R^{n_{in}}$ and a target $\cv_s \in \R^{n_{out}}$. 
Given a dataset~$\D$, the idea behind supervised learning is to construct a model~${f_m: \R^{n_{in}} \rightarrow \R^{n_{out}}}$, which captures the relationship between input and target.
The model~$f_m$ typically has the following form:
\begin{align}
f_m(\xv) := \pazocal{P}(\Wm_K f_p (\xv) + \bv_K), 
\label{eq:for_map}
\end{align}
where~$\pazocal{P}: \R^{n_{out}} \rightarrow \R^{n_{out}} $ is a hypothesis function and $f_p: \R^{n_{in}} \rightarrow \R^{n_{fp}}$ denotes a nonlinear feature extractor, often called forward propagation.
The parameters~$\Wm_K \in \R^{n_{out} \times {n_{fp}}}$ and $\bv_K \in \R^{n_{out}}$ are used to perform an affine transformation of the extracted features. 
Through the manuscript, we often denote~$\Wm_K$ and $\bv_K$ collectively as~${\thetav_K:=( \text{flat}(\Wm_K), \text{flat}(\bv_K))}$, where the function $\text{flat}(\cdot)$ is used to convert a tensor into a 1-dimensional array.

This work builds upon a continuous-in-depth approach~\cite{queiruga2020continuous, chang2017multi, weinan2017proposal}, which 
interprets the forward propagation through the network as a discretization of the nonlinear ordinary differential equation (ODE). 
Thus, let us consider the following dynamical system:
\begin{equation}
\begin{aligned}
\partial_t \vec{q}(t) &= \pazocal{F}(\vec{q}(t), \vec{\theta}(t)), & \quad \forall t \in (0,T), \\
\vec{q}(0) &= \mat{Q} \xv, & 
\end{aligned}	
\label{eq:dynamical_system}
\end{equation}
where~$\vec{q}(t):\R \rightarrow \R^{n_{fp}}$ and $ \vec{\theta}(t): \R \rightarrow \R^{n_{c}}$ denote time-dependent state and control functions, respectively. 
Here, the symbol $n_c$ denotes the size of controls associated with a given time $t$.
The system~\eqref{eq:dynamical_system} continuously transforms input features~$\vec{x}$ into the final state~$\vec{q}(T)$, defined at the time $T$.
The initial condition in~\eqref{eq:dynamical_system} is used to map an input~$\vec{x}$ into the dimension of the system's dynamics, denoted by~$n_{fp}$.
This is achieved using the linear operator~$\mat{Q} \in \R^{{n_{fp}} \times n_{in}}$, which can be defined apriori or learned during the training process. 
The right-hand side, function~${\pazocal{F}: \R^{n_{fp}} \times \R^{n_{c}} \rightarrow \R^{n_{fp}}}$, is often called a residual block. 
An exact form of the function~${\pazocal{F}}$ is typically prescribed by the network architecture. 
For instance, it can be a single layer perceptron or a stack of multiple convolutional layers. 
Note, the function~$\pazocal{F}$ has to fulfill certain assumptions, e.g., the Lipschitz-continuity, so that solution of~\eqref{eq:dynamical_system} exists, see~\cite{clarke2005maximum} for details.

\begin{remark}
Formulation~\eqref{eq:dynamical_system} gives rise to ResNets with a constant width. 
More practical scenarios will be considered in~\cref{sec:ml_hierarchy_resnet}. 
\end{remark}

Finally, we can formulate the supervised learning problem as a continuous optimal control problem~\cite{haber2017stable}, thus as
\begin{align}
  &\underset{\vec{\theta}, \vec{q} }{\text{min}} \ \ \   \frac{1}{n_{s}} \sum_{s=1}^{n_{s}} \ell(\underbrace{ \pazocal{P}(\Wm_K \vec{q}_s(T) + \bv_K)}_{ \vec{y}_s }, \cv_s) + \frac{\beta_1}{2} \int\limits_{0}^{T} \pazocal{R}(\vec{\theta}(t) ) \ dt +   \frac{\beta_2}{2} \pazocal{S}(\vec{\theta}_K) ,\nonumber \\  
  & \text{subject to}  \quad \partial_t \vec{q}_s(t) = \pazocal{F}(\vec{q}_s(t), \vec{\theta}(t)),   \qquad  \forall t \in (0,T),  \label{eq:cts_problem}  \\
& \quad \quad \quad \quad \quad \quad  \vec{q}_s(0) = \mat{Q}\xv_s, \nonumber
\end{align}
where~$\vec{q}_s(T) \in \R^{n_{fp}}$ is the output of the dynamical system~\eqref{eq:dynamical_system} for a given sample~$\xv_s$. 
The symbols $\pazocal{R}, \pazocal{S}$ and $\beta_1, \beta_2 > 0$ denote convex regularizers and their parameters, respectively. 
A loss function ${\ell: \R^{n_{out}} \times \R^{n_{out}} \rightarrow \R}$ measures the deviation of the predicted output $ \vec{y}_s \in \R^{n_{out}}$, given as $ \vec{y}_s:=\pazocal{P}(\Wm_K \vec{q}_s(T) + \bv_K) $, from the target~$\cv_s$.
An exact form of loss function depends on the problem at hand. 
In this work, we use least squares and cross-entropy loss functions~\cite{goodfellow2016deep} for regression and classification tasks, respectively.

\subsection{Discrete minimization problem}
In order to solve the minimization problem~\eqref{eq:cts_problem} numerically, we discretize the temporal domain into $K-1$ uniformly distributed time intervals. 
Thus, we consider the time-grid ${0 = \tau_0 < \cdots < \tau_{K-1} = T}$ of $K$ uniformly distributed time points.
Given a uniform time-step ${\Delta_t: = T/(K-1)}$, the k-th time point is defined as~${\tau_k:= \Delta_t k}$.
Now, states and controls can be approximated at a given time~$\tau_k$ as~$\vec{q}_k \approx \vec{q}(\tau_k)$, and $\vec{\theta}_k \approx \vec{\theta}(\tau_k)$, respectively.

To construct state approximations, one can utilize a numerical integration scheme. 
Here, we employ the explicit (forward) Euler scheme, as it is simple and computationally efficient.
However, more stable integration schemes can be employed.
We refer the interested reader to~\cite{hirsch1974differential} for an overview of various integration schemes.
Note, the stability of the explicit Euler scheme can be ensured by employing a sufficiently small time-step~$\Delta_t$. 

The approximation of controls at $\tau_k$ can be obtained as $\vec{\theta}_k(\tau_k) = \sum_{k=0}^{K-1} \vec{\theta}_k \phi_k(\tau_k)$,
where each coefficient $\vec{\theta}_k $ is associated with the k-th node of the time-grid. 
Here, we employ piecewise-constant basis functions, defined as 
\begin{align}
 \phi_k(t) = 
 \begin{cases}
 1, \quad t \in [k \Delta t , (k+1) \Delta t), \\
 0, \quad \text{otherwise},
 \end{cases}
 \label{eq:basis}
\end{align}
for all $k =0, \ldots, K-1$. 
Altogether, this gives rise to a network with~$K$ layers and imposes a standard ResNet architecture with identity skip connections~\cite{he2016identity}. 
Each k-th layer is then associated with a state $\vec{q}_k$ and controls/parameters $\vec{\theta}_k$. 
We note that alternative approaches, where controls and states are decoupled across layers, were recently also considered in the literature, see for instance~\cite{queiruga2020continuous, gunther2021spline, massaroli2020dissecting}.

Now, we can obtain the following discrete minimization problem:
\begin{align}
 & \underset{\vec{\theta}, \vec{q}}{\text{min}} \ \tilde{\L}(\vec{\theta}, \vec{q}) := \ \frac{1}{n_s} \sum_{s=1}^{n_s} \ell(\yv_s, \vec{c}_s)  + \frac{\beta_1}{2} \sum_{k=1}^{K-1} \pazocal{R}(\vec{\theta}_{k-1}, \vec{\theta}_{k}) + \frac{\beta_2}{2} \pazocal{S}(\vec{\theta}_K), \nonumber \\
 &\text{subject to } \ \ \vec{q}_{s, k+1} = \ \vec{q}_{s, k} + \Delta_t \pazocal{F}(\vec{q}_{s, k}, \vec{\theta}_k), \qquad \forall k=0, \ldots, K-1, \label{eq:discrete_problem1} \\
& \qquad \qquad \qquad \ \vec{q}_{s, 0} = \ \mat{Q} \vec{x}_s, \qquad \qquad \qquad \qquad \quad \ \forall s=1, \ldots, n_s,  \nonumber
\end{align}
where $\vec{q}_{s, k}$ denotes the state associated with the $s$-th sample and the $k$-th layer. 
The symbol~$\vec{\theta} \in \R^n$ is used to collectively denote all the network parameters, i.e., ${\vec{\theta}=( \text{flat}(\vec{\theta}_0), \dots, \text{flat}(\vec{\theta}_{K-1}), \text{flat}(\vec{\theta}_K))}$. 
For all layers $k=1, \ldots, K-1$, we employ the following regularizer:~$\pazocal{R}(\vec{\theta}_{k-1}, \vec{\theta}_{k}) := \frac{1}{2 \Delta t} \| \vec{\theta}_k - \vec{\theta}_{k-1} \|^2 $, which ensures that the parameters vary smoothly across adjacent layers. 
In addition, we regularize the parameters of the hypothesis function as~$\pazocal{S}(\vec{\theta}_K) := \frac{1}{2} \| \Wm_K \|^2_F + \frac{1}{2} \| \bv_K \|^2$, where $ \| \cdot \|_F^2$ denotes the Frobenius norm.

Instead of solving the equality constrained minimization problem~\eqref{eq:discrete_problem1} directly, we can eliminate dependence on~$\vec{q}$ by time-stepping (forward propagation) and focus only on solving the following reduced unconstrained minimization problem: 
\begin{align}
 \label{eq:discrete_problem}  
 \underset{\vec{\theta} \in \R^{n}}{\text{min}} \ {\L}(\vec{\theta}) = \tilde{\L}(\vec{\theta}, \vec{\tilde{q}}).
\end{align}
Here, the states~$\vec{\tilde{q}}$ are obtained by explicitly satisfying the constraint in~\eqref{eq:discrete_problem1}, for given parameters~$\vec{\theta}$. 
Solving the minimization problem~\eqref{eq:discrete_problem} is called training.
Usually, the training is performed using a first-order optimizer, which requires knowledge of the reduced gradient~$\nabla_{\vec{\theta}} \L$. 
In this work, we obtain~$\nabla_{\vec{\theta}} \L$ using the back-propagation technique~\cite{chauvin1995backpropagation}, efficient implementation of which is provided in various deep-learning frameworks.
Please note, the dynamic in~\eqref{eq:discrete_problem} is decoupled across the samples.
Therefore, the reduced gradient can be evaluated using only a portion of the dataset~$\pazocal{D}$.
This is often utilized by stochastic/mini-batch solution strategies, such as stochastic gradient descent (SGD)~\cite{robbins1951stochastic} or Adam~\cite{adam2014}.

\section{Globally convergent multilevel training}
\label{sec:rmtr_resnet}
In this work, we propose to minimize the discrete optimization problem~\eqref{eq:discrete_problem} using a variant of the RMTR method~\cite{Gratton2008recursive, Gross2009}. 
The RMTR method incorporates the trust-region globalization strategy into the nonlinear multilevel framework, which gives rise to a globally convergent nonlinear multilevel method. 
The method was originally proposed for solving deterministic convex/non-convex minimization problems arising from the discretization of PDEs. In this section, we propose to extend the applicability of the RMTR method to the training of ResNets. 
We briefly describe the algorithm and discuss how to construct the multilevel hierarchy and transfer operators by taking into account the structure of the underlying optimization problem~\eqref{eq:discrete_problem}. 
An extension of the method into stochastic settings will be carried out in \cref{sec:dynamic_sampling}.

\begin{figure}[t]
	\centering
\scalebox{.75}{		
\includegraphics{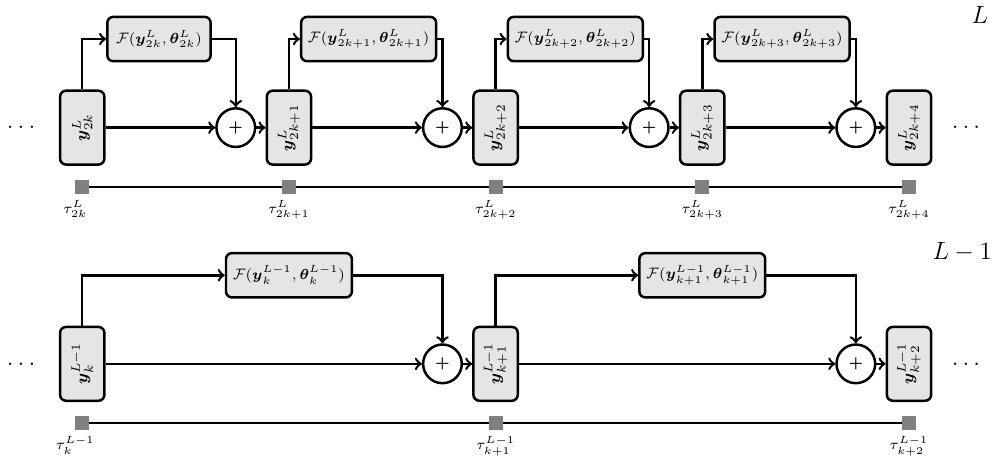}
    }
    \caption{An example of a multilevel hierarchy of ResNets. The state and control variables are discretized using different time grids.}
    \label{fig:ml_resnets}
\end{figure}

\subsection{Multilevel framework}
\label{sec:ml_hierarchy_resnet}
The proposed multilevel training method relies on a hierarchy of~$L$ levels, denoted by~$l = 1, \dots, L$. 
Each level~$l$ is associated with a ResNet of different depth, obtained by discretizing continuous optimal control problem~\eqref{eq:cts_problem}. 
More precisely, we discretize the time interval~$(0, T)$ on the coarsest level, $l=1$, with some prescribed time-step~$\Delta_t^{1}$. 
On all other levels, we use time-step~$\Delta_t^{l}$, obtained as~$\Delta_t^{l}=0.5\Delta_t^{l-1}$. 
The multilevel hierarchy of ResNets obtained in this way is then used internally by the RMTR method to accelerate the training of the ResNet on the finest level. 
Since we employ a uniform refinement in time by a factor of two, the number of layers and parameters is doubled between two subsequent levels.
As a consequence, it is roughly two--times more expensive to perform one forward-backward propagation on level~$l+1$ than on level~$l$.
\Cref{fig:ml_resnets} illustrates a two-level hierarchy of ResNets.

\subsubsection{Transfer operators}
\label{sec:transfer}
The transfer of the data between different levels of the multilevel hierarchy is performed using transfer operators. 
The prolongation operator $\Pm_{l}^{l+1} \in \R^{n^{l+1} \times n^{l}}$ is used to transfer quantities, such as search-directions, from level $l$ to level $l+1$. 
We assemble the prolongation operator $\Pm_{l}^{l+1}$ in a two-step process. 
Firstly, we simply copy the parameters contained in~$\Qm^l$ and~$\vec{\theta}_K^l$ from level $l$ to the level $l+1$. 
Thus, the prolongation operator is the identity, since $\Qm^l$, and~$\vec{\theta}_K^l$ are represented by ResNets on all levels. 
In the second step, we prolongate the network parameters obtained by the discretization of the dynamical system, recall~\cref{sec:resnet}. 
Here, we make use of the fact that we can change the basis functions used for evaluating~$\vec{\theta}^{l}(\tau_k)$ by projecting to a refined basis.  
For example, the network parameters can be prolongated as follows:
\begin{equation}
\begin{aligned}
\vec{\theta}^{l+1}_{2k} &= \vec{\theta}^{l}_k, \quad \qquad \text{and} \quad \qquad
\vec{\theta}^{l+1}_{2k+1} &= \vec{\theta}^{l}_k, \quad \qquad \forall k \in 0, \ldots, K^l-1, 
\end{aligned}
\end{equation}
if the piecewise constant basis functions, defined in~\eqref{eq:basis}, are employed. 
Here, the symbol~$K^l$ denotes number of layers associated with ResNet on level $l$. 
This type of prolongation is well known in multigrid literature as piecewise constant interpolation.
In the context of ResNets, this particular type of transfer operator was employed for the first time in~\cite{haber2018learning, chang2017multi} for the cascadic SGD training.

Furthermore, we also employ the restriction operator~${\Rm_{l+1}^{l} \in \R^{n^{l} \times n^{l+1}}}$ to transfer the gradients  from level $l+1$ to level $l$.
As common in practice, the operator $\Rm_{l+1}^{l} $ is assumed to be the adjoint of~$\Pm_{l}^{l+1}$, i.e.,~$\Rm_{l+1}^{l} = (\Pm_{l}^{l+1})^T$.
We also use operator $\boldsymbol{\Pi}_{l+1}^{l} \in \R^{n^{l} \times n^{l+1}}$ to transfer the network parameters from level $l+1$ to level $l$.
The operator $\boldsymbol{\Pi}_{l+1}^{l}$ is obtained by scaling the restriction operator $\Rm_{l+1}^{l}$, thus as $\boldsymbol{\Pi}_{l+1}^{l} := \Dm \Rm_{l+1}^{l}$, 
where $\Dm \in \R^{{n^{l+1}} \times n^{l+1}}$ is a diagonal matrix. 
The diagonal of $\Dm$ takes on the value $1$ for the rows associated with the transfer of parameters~$\Qm^{l+1}$, and~$\vec{\theta}_K^{l+1}$ and the value $0.5$ otherwise. 
The use of scaling matrix~$\Dm$ ensures that the magnitude of network parameters does not grow on the coarser levels, which would lead to numerical instabilities, such as exploding gradients.
More details regarding the choice of projection operator $\boldsymbol{\Pi}_{l+1}^{l}$ can be found in the supplement, \cref{sec:supl_rmtr_params}.

\subsubsection{Networks with varying width}
\label{sec:nets_with_var_width}
Until now, we considered only ResNets with constant width.
This is due to the fact that dynamical systems, such as~\eqref{eq:dynamical_system}, do not allow for a change of dimensionality. 
The projection of the dynamics to a space of higher/lower dimension can be performed only at time~$t=0$ or~$t=T$. 
However, deep learning practitioners quite often utilize networks with varying width. 
Special interest is put into convolutional networks, which split a network into~$A$ stages. 
Each stage is then associated with a different network width (number of filters), and image resolution. 
The change in dimensionality between different stages is usually performed by downsampling~\cite{goodfellow2016deep}. 

We can incorporate~$A$-stage network architectures into our multilevel framework by interpreting their forward propagation as a composition of several dynamical systems~\cite{queiruga2020continuous}. 
The~$A$-stage network is then obtained by stitching together~$A$ dynamical systems as follows:
\begin{equation}
\begin{aligned}
\partial \vec{q}_{a}(t) &= \pazocal{F}(\vec{q}_{a}(t), \vec{\theta}_{a}(t)), \qquad \quad \ \   \forall t \in (0, T_a), \quad \forall {a} \in \{ 1, \ldots, A \}, \\ 
\vec{q}_{a}(0) &= 
\begin{cases}
\Qm_{a} \xv, \qquad \qquad \qquad \ \ &\text{if} \ {a} =1, \\
\Qm_{a} \vec{q}_{{a}-1}(T_{{{a}}-1}), \qquad \ \ &\text{otherwise}.
\end{cases}
\end{aligned}
\label{eq:cont_stages}
\end{equation}
Thus, the {a}-th stage is associated with a dynamical system, which transforms the input~$\xv$ or the output of the previous stage~$\vec{q}_{{a}-1}(T_{{a}-1})$ into~$\vec{q}_{{a}}(0)$. 
The matrices~$\{ \Qm_{{a}} \}_{{a}=1}^{{A}}$, in~\eqref{eq:cont_stages} incorporate the dimensionality change {between different stages}.
{Note, that the size of matrices~$\{ \Qm_{{a}} \}_{{a}=1}^{{A}}$ varies. 
In particular, $\Qm_{1} \in \R^{fp_{1} \times n_{in}}$, while $\Qm_{a} \in \R^{fp_{a} \times fp_{a-1}}$ for all $a>1$, where $fp_{a}$ denotes the network width of the a-th stage.}

\begin{figure}[t]
	\centering
\scalebox{.7}{	
\includegraphics{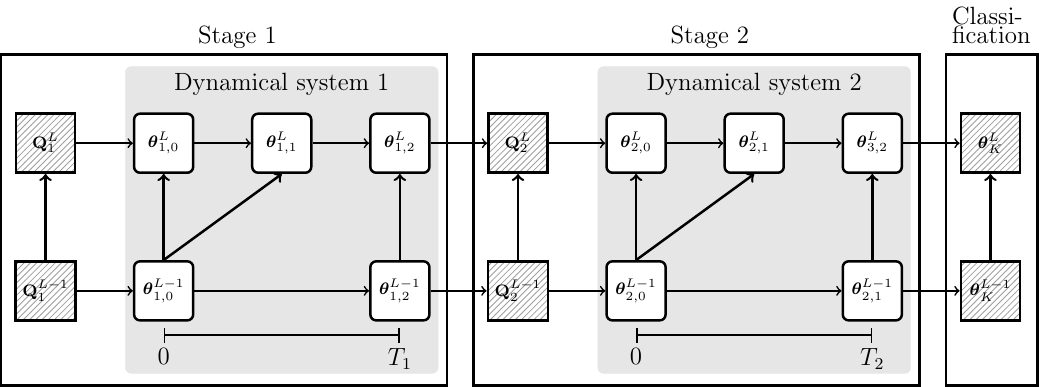}	
    }
    \caption{{
 An assembly process of the prolongation operator for 2-stage ResNet.
 In the first step (hatched squares), we copy $\Qm_1^{L-1}, \Qm_2^{L-1}$ and $\vec{\theta_K}^{L-1}$ to the finer level.
 In the second step (rounded white squares), we prolongate parameters associated with the discretization of the dynamical systems.
 Note, the description of the parameters uses double subscript to denote the stage and the layer index. 
 For simplicity, the illustration does not visualize the skip connections.
    }}
    \label{fig:ml_resnets_transfer}
\end{figure}

Since our goal is to obtain the standard ResNet architecture~\cite{he2016deep}, we can again discretize all time derivatives in~\eqref{eq:cont_stages} using the explicit Euler method. 
Similarly to the previous section, we can obtain a multilevel hierarchy of~{{$A$}-stage} ResNets by discretizing the dynamical systems~\eqref{eq:cont_stages} with varying discretization parameters. 
The construction of transfer operators also follows the discussion from \cref{sec:transfer}.
{Here, we highlight the fact that the transfer of the parameters is always performed only within a given stage, i.e., the parameters are never transferred across multiple stages, {see also \cref{fig:ml_resnets_transfer}}. }

\begin{remark}
The choice of the time interval~$(0, T_{{a}})$ and the discretization strategy associated with different dynamical systems in~\eqref{eq:cont_stages} can differ.
\end{remark}

\subsection{The RMTR method}
\label{sec:rmtr_algo}
In this section, we provide a brief description of the RMTR method~\cite{Gratton2008recursive, Gross2009}. 
We describe the RMTR algorithm in {the} form of a V-cycle, but other cycling schemes, such as F-cycle, can also be used in practice.  
{Throughout} this section, we use superscript and subscript to denote the level and iteration index, respectively. 
For instance, the symbol $\vec{\theta}_i^l$ denotes the network parameters associated with level~$l$ and iterate~$i$.

As common for {the} nonlinear multilevel methods, such as FAS~\cite{brandt1977multi}, or MG/OPT~\cite{Nash2000multigrid}, the RMTR method approximates~\eqref{eq:discrete_problem} on each level $l$ by means of some level-dependent objective function $\H^l:\R^{n^l} \rightarrow \R$. 
In this work, we assume that a function~$\H^l$ is computationally less expensive to minimize than $\H^{l+1}$, and that $n^l < n^{l+1}$ for all $l=1, \ldots, L-1$.
On the finest level, we define~$\H^L$ as~${\H^L := \L^L}$, thus $\H^L$ denotes an objective function of the minimization problem at the hand. 
On coarser levels, we aim to construct the function~$\H^l$ such that its (approximate) minimization yields a good search-direction on the level $l+1$. 
Here, we construct~$\{\H^l \}_{l=1}^{L-1}$ using knowledge of the loss functions $\{\L^l \}_{l=1}^{L-1}$, obtained by discretizing~\eqref{eq:cts_problem} with different discretization parameters.

The V-cycle of the RMTR method starts on the finest level, $l=L$, with some initial parameters $\vec{\theta}_0^L$. 
The algorithm then passes through all levels of the multilevel hierarchy until the coarsest level, $l=1$, is reached. 
On each level $l$, we perform a pre-smoothing step to improve the current iterate, i.e., parameters $\vec{\theta}_0^l$. 
The smoothing step is carried out using $\mu_s$ iterations of the trust-region method~\cite{Conn2000trust}.
The trust-region method produces the sequence of the search-directions $\{\sv_i^l \}$ by (approximately) minimizing the following trust-region subproblem:
\begin{equation}
\begin{aligned}
&\underset{\sv_i^l \in \R^{n^l}}{\text{min}} \ m_i^l(\vec{\theta}_i + \sv_i^l ) := \H^l(\vec{\theta}_i^l) + \langle \nabla \H^l(\vec{\theta}_i^l), \sv_i^l \rangle + \frac{1}{2} \langle \sv_i^l,  \Bm^l_i \ \sv^l_i \rangle,  \\
&\text{subject to} \ \ \| \sv_i^l \| \leq \Delta_i^l,
\end{aligned}
\label{eq:tr_subproblem}
\end{equation}
where model $m_i$ is constructed as a second-order Taylor approximation of $\H^l$ around current iterate~$\vec{\theta}_i^l$. 
The model $m_i$ does not employ the exact Hessian $\nabla^2 \H^l(\vec{\theta}_i^l)$, but rather its approximation~$ \Bm^l_i \approx \nabla^2 \H^l(\vec{\theta}_i^l)$. 
This is due to the fact that the assembly of the exact Hessian~$\nabla^2 \H^l(\vec{\theta}_i^l)$ is often prohibitive in practice, especially for very deep networks, by virtue of memory requirements. 
In this work, we approximate~$\nabla^2 \H^l(\vec{\theta}_i^l)$ using the limited memory SR1 (L-SR1) method~\cite{nocedal1980updating, nocedal2006numerical}, which utilizes a recursive rank-one update formula. 
We remark that it is also possible to use only first-order information by simply setting $ \Bm^l_i$ to identity, thus as~$ \Bm^l_i=\Im$.

The trust-region method enforces convergence control in two steps. 
Firstly, the constraint in~\eqref{eq:tr_subproblem} ensures that the size of the search-direction $\sv_i^l$ is bounded by the trust-region radius~$\Delta_i^l > 0$. 
Secondly,  the search-direction $\sv_i^l$, obtained by solving~\eqref{eq:tr_subproblem}, is accepted by the algorithm only if $\rho_i^l > \eta_1$, where $\eta_1>0$, and
$ \rho_i^l$ is given as
\begin{align}
 \rho_i^l =  \frac{\H^l(\vec{\theta}_i) - \H^l(\vec{\theta}_i + \sv^l_i)}{m^l(\vec{\theta}_i) -  m^l(\vec{\theta}_i + \sv^l_i)}.
\end{align}
The trust-region ratio~$\rho_i^l$ is also employed to adjust the size of the trust-region radius.
In particular, the trust-region radius is adapted as outlined in~\cref{alg:conv_control}.

After the pre-smoothing step is performed, the algorithm advances to a subsequent coarser level. 
To this aim, the parameters and trust-region radius are initialized as~$\vec{\theta}_{0}^{l-1} := {\boldsymbol{\Pi}^{l-1}_l} \vec{\theta}_{\mu_s}^l$, and~$\Delta^{l-1}_0 := \Delta^l_{\mu_s}$, respectively. 
We also construct the coarse-level objective function~$\H^{l-1}$ using the knowledge about the current fine level iterate~$\vec{\theta}_{\mu_s}^l$ and the gradient~$\nabla \H^l(\vec{\theta}_{\mu_s}^l)$. 
More precisely, the function $\H^{l-1}$ is constructed as 
\begin{align}
\H^{l-1}(\vec{\theta}^{l-1}_0 + \sv_i^{l-1}) := \L^{l-1}(\vec{\theta}^{l-1}_0 + \sv_i^{l-1}) + \langle \delta \gv^{l-1}, \sv_i^{l-1} \rangle,
\label{eq:coarse_objective_galerkin}
\end{align}
where $\delta \gv^{l-1}:= \Rm^{l-1}_{l} \nabla \H^{l}(\vec{\theta}^{l}_{\mu_s}) - \nabla \L^{l-1}(\vec{\theta}^{l-1}_{0})$. 
The linear term $\delta \gv^{l-1}$ takes into account the difference between restricted fine-level gradient~$\Rm^{l-1}_{l} \nabla \H^{l}$ and the initial gradient of the coarse-level loss function~$\nabla \L^{l-1}(\vec{\theta}^{l-1}_{0})$.
The presence of this term ensures that the first-step of the minimization process on level $l-1$ is performed in the direction of the negative restricted fine-level gradient, thus that~${\nabla \H^{l-1}(\vec{\theta}^{l-1}_0) = \Rm^{l-1}_{l} \nabla \H^{l} (\vec{\theta}_{\mu_s}^l)}$.

The aforementioned process is repeated until the coarsest level, $l=1$, is reached. 
Once the coarsest level is entered, the algorithm carries out~$\mu_c$ iterations of the trust-region method and produces the updated parameters $\vec{\theta}_{\mu_c}^1$.
Subsequently, the algorithm returns to the finest level by transfering the correction obtained on the level $l$, i.e.,~$\vec{\theta}_{\mu^{l}}^{l} - \vec{\theta}_{0}^{l}$, to the level~$l+1$. 
Here, we use the symbol~$\mu^{l}$ to collectively denote all iterations taken on level $l$. 
As common for {the} trust-region based methods, the quality of the prolongated coarse-level correction $\sv_{\mu_s+1}^{l+1} := \Pm^{l+1}_{l}(\vec{\theta}_{\mu^{l}}^{l} - \vec{\theta}_{0}^{l})$ has to be assessed before it is accepted by the level $l+1$.  
To this aim, we employ a multilevel TR ratio, defined as 
\begin{align}
\rho^{l+1}_{\mu_s+1} := \frac{\H^{l+1}(\vec{\theta}_{\mu_s}^{l+1}) - \H^{l+1}(\vec{\theta}_{\mu_s}^{l+1} + \sv_{\mu_s+1}^{l+1})}{\H^{l}(\vec{\theta}_{0}^{l}) - \H^{l}(\vec{\theta}_{\mu^{l}}^{l})}.
\label{eq:rho_ml}
\end{align}
If $\rho^{l+1}_{\mu_s+1} > \eta_1$, then it is safe to accept the prolongated coarse-level correction $\sv_{\mu_s+1}^{l+1}$. 
Otherwise, $\sv_{\mu_s+1}^{l+1}$ has to be disposed. 
Additionally, the TR radius has to be updated accordingly. 
This can be achieved by utilizing the update rules outlined in \cref{alg:conv_control}. 
In the end, the RMTR algorithm performs $\mu_s$ post-smoothing steps at a given level $l$.
This process is repeated on every level until the finest level is reached. 
\Cref{alg:rmtr_algo} summarizes the described process (V-cycle of the RMTR method).

\begin{algorithm}[t]
\caption{\footnotesize RMTR($l, \  \H^l, \  \vec{\theta}^l_{0},  \  \Delta^l_{0}$)}
\label{alg:rmtr_algo}
\begin{algorithmic}[1]
\footnotesize
\Require{\footnotesize $ l \in \N, \  \H^l:\R^{n^l}\rightarrow \R,  \ \vec{\theta}^l_{0} \in \R^{n^l}, \  \Delta^l_{0} \in \R$}
\Constants {\footnotesize  $\mu_s, \ \mu_c \in \N$}
\State \footnotesize  $ [\vec{\theta}_{\mu_s}^l, \  \Delta_{\mu_s}^l] = \text{TrustRegion}(\H^l,  \ \vec{\theta}^l_{0},  \  \Delta_{0}^l, \  \mu_s )$
\Comment{\footnotesize Pre-smoothing}
\State \footnotesize Construct~$\H^{l-1}$  \Comment{\footnotesize Initialize coarse-level objective function}
\If{\footnotesize $l ==2$}
\State \footnotesize  $ [\vec{\theta}_{\mu^{l-1}}^{{l-1}}] = \text{TrustRegion}(\H^{l-1},  \ {\boldsymbol{\Pi}^{l-1}_l} \vec{\theta}_{\mu_s}^l,  \  \Delta_{\mu_s}^{l}, \  \mu_c )$  \Comment{\footnotesize Coarse-level solve}
\Else
\State \footnotesize $[\vec{\theta}_{\mu^{l-1}}^{l-1}]$ = RMTR($l-1, \H^{l-1}, {\boldsymbol{\Pi}^{l-1}_l} \vec{\theta}_{\mu_s}^l, \ \Delta_{\mu_s}^l) $  \Comment{\footnotesize Call RMTR recursively}
\EndIf
\State \footnotesize $\sv^l_{\mu_s+1} = {\Pm}_{l-1}^{l}(\vec{\theta}_{\mu^{l-1}}^{l-1} - {\boldsymbol{\Pi}^{l-1}_l} \vec{\theta}_{\mu_s}^l)$  \Comment{\footnotesize Prolongate coarse-level correction}
\State \footnotesize Compute $\rho^l_{\mu_s +1}$ by means of~\eqref{eq:rho_ml}
\State \footnotesize $ [\vec{\theta}_{\mu_s+1}^l, \  \Delta_{\mu_s+1}^l]$ = ConvControl($\rho^l_{\mu_s +1}, \ \vec{\theta}_{\mu_s}^l, \ \sv^l_{\mu_s+1}, \ \Delta_{\mu_s}^l$) 
 \Comment{\footnotesize Call \cref{alg:conv_control}}
\State \footnotesize $ [\vec{\theta}_{\mu^l}^l, \  \Delta_{\mu^l}^l] = \text{TrustRegion}(\H^l,  \ \vec{\theta}_{\mu_s+1}^l, \  \Delta_{\mu_s+1}^l, \  \mu_s )$
 \Comment{ \footnotesize Post-smoothing}
\State \Return \footnotesize $\vec{\theta}_{\mu^l}^l$, $\Delta_{\mu^l}^l$ 
\end{algorithmic}
\end{algorithm}

 \section{Multilevel training - hybrid (stochastic-deterministic) settings}
\label{sec:dynamic_sampling}
The nonlinear minimization problem~\eqref{eq:discrete_problem} is non-convex, hence its minimization admits multiple local minimizers. 
We aim to find a solution, i.e.,~a set of parameters, which generalizes well to previously unseen examples. 
It has been observed in practice, that flat minimizers generalize better than sharp minimizers~\cite{keskar2016large, golmant2018computational, hoffer2017train}. 
The study provided in~\cite{keskar2016large} demonstrates that the large-batch/deterministic methods tend to be attracted to sharp minimizers.
Instead, small-batch methods tend to be more exploratory, which helps them to escape basins of attraction of sharp minimizers and converge to flat minimizers.
However, there are practical reasons why large-batch methods should be employed.
For example, they enable faster convergence in the local neighborhood of a minimizer~\cite{keskar2016large}. 
Moreover, large-batch methods use computational resources more efficiently, e.g., by decreasing data movement between a CPU and a GPU device.

In this work, we take advantage of both small-batch and large-batch techniques by using the RMTR method in conjunction with the dynamic sample size (DSS) strategy. 
This gives rise to the hybrid stochastic-deterministic multilevel method, named dynamic sample sizes RMTR (DSS-RMTR) method. 
The DSS-RMTR starts the training process in a stochastic regime, which uses only a small subset of all samples, called mini-batch, in order to evaluate an objective function and its gradient.
As training progresses, the objective function and gradient are evaluated with increasing accuracy, i.e.,~by considering a larger subset of all samples. 
Eventually, the full dataset is used and the {DSS-RMTR} method operates in the deterministic regime. 
At this point, the global convergence properties of the DSS-RMTR method follow directly from the theory developed in~\cite{Gratton2008recursive, Gross2009}.

Similarly to the adaptive sample size trust-region (ASTR) method~\cite{mohr2019adaptive}, the DSS-RMTR adjusts mini-batch sizes using information about the objective function evaluated using the full dataset~$\D$. 
The DSS-RMTR method differs from ASTR in two main aspects. 
Firstly, the search-direction associated with a given mini-batch is obtained using a V-cycle of the RMTR method, not 
an iteration of the single-level trust-region method. 
Secondly, the DSS-RMTR method incorporates the curvature information by means of limited-memory secant methods. 
In contrast, the numerical results presented in~\cite{mohr2019adaptive} rely only on first-order information. 
We remark that using limited-memory secant methods, such as L-SR1, within the stochastic regime is not trivial and requires several adjustments compared to the deterministic regime~\cite{berahas2020robust, berahas2019quasi, erway2020trust}.

\subsection{DSS-RMTR algorithm}
The DSS-RMTR algorithm consists of two phases: global and local/mini-batch. 
The global phase is performed using a full dataset~$\D$, while the local phase utilizes subsets of dataset~$\D$. 
Through the following, we use the subscript pair~$(e,b)$ to denote quantities associated with global and local phases, e.g.,~$\vec{\theta}_{e,b}$ denotes parameters obtained during~$e$-th epoch using mini-batch~$b$. 
Since the dynamic sampling strategy acts only on the finest level, our description omits superscripts specifying a given level.

\subsubsection{Local phase}
The local phase starts by generating a set of mini-batches~$\{ \D_b \}_{b=1}^{n_e}$, where~${n_e \geq 1}$. 
Samples of each mini-batch~$\D_b$ are extracted from the dataset~$\D$, such that each~$\D_b$ contains~$\text{mbs}_e$ samples. 
 Once the mini-batches $\{ \D_b \}_{b=1}^{n_e}$ are created, we construct a set of local optimization problems. 
Each local optimization problem has the same form as the minimization problem~\eqref{eq:discrete_problem}, but the loss function is evaluated using only samples from one mini-batch.
We denote the sub-sampled objective functions associated with local optimization problems collectively as~$\{ \L_b \}_{b=1}^{n_e}$. 
These local optimization problems are then approximately solved, using one V-cycle of the RMTR method, in a successive manner. 
Thus, the parameters~$\vec{\theta}_{e,b}$, obtained by minimizing~$\L_b$, are used as an initial guess for the minimization of the function~$\L_{b+1}$.  
A local phase terminates once we have iterated over all mini-batches. 

\subsubsection{Global phase}
In a global phase, the DSS-RMTR method determines the quality of a trial point~$\vec{\theta}_{e, n_e}$, obtained as a result of the local phase. 
This is achieved by using global trust-region ratio~$\rho^G_e$, defined as
\begin{align*}
\rho^G_e = \frac{\L(\vec{\theta}_{e,0}) - \L(\vec{\theta}_{e,\text{n}_e}) }{\frac{1}{\text{n}_e} \sum_{b=1}^{\text{n}_e} \big(\L_b(\vec{\theta}_{e,b}) - \L_b(\vec{\theta}_{e,b+1}) \big)} = \frac{\text{global reduction}}{\text{average local reduction}}.
\end{align*}
Thus, the global trust-region ratio~$\rho^G_e$ compares the actual reduction observed in the global objective function~$\L$ and an average local reduction, obtained while minimizing the local objective functions~$\{ \L_b \}_{b=1}^{n_e}$. 

As customary for trust-region algorithms, the trial point~$\vec{\theta}_{e, n_e}$ is accepted only if~${\rho^G_e > \zeta_1}$, where~${\zeta_1 > 0}$. Otherwise, we reject the trial point.
In addition, the global trust-region ratio~$\rho_e^G$ is used to adjust the mini-batch size.
Since small values of~$\rho_e^G$ indicate that~$\{ \L_i \}_{i=1}^{n_e}$ do not approximate~$\L$ well, we increase the mini-batch size. 
Thus, we decrease the number of mini-batches, but each mini-batch will contain a larger portion of samples from~$\D$, i.e.,~$\text{mbs}_{e+1} > \text{mbs}_{e}$.
In contrast, large values of~$\rho_e^G$ suggest that the averaged sub-sampled objective functions~$\{ \L_b \}_{b=1}^{n_e}$ approximate~$\L$ well and can be used during the next epoch. 
The described process is summarized in \cref{alg:DSSRMTR}.

\begin{remark}
Numerical evaluation of the global trust-region ratio~$\rho^G_e$ is an expensive operation, especially if the number of samples in the dataset~$\D$ is large.
We can decrease the computational cost by performing the local phase multiple times before a global phase takes place.
\end{remark}

\subsubsection{{Properties of DSS-RMTR algorithm}}
{In this section, we comment on the convergence properties and the practical implementation of the proposed DSS-RMTR method. }

\paragraph{{Convergence}}
{The DSS-RMTR method intertwines RMTR method~\cite{Gratton2008recursive} with DSS strategy~\cite{mohr2019adaptive}.
Authors of~\cite{mohr2019adaptive} show theoretically that after a finite number of epochs, the mini-batch size is increased by DSS strategy sufficiently many times, such that it coincides with the size of the full dataset. 
At this point, the minimization of~\eqref{eq:discrete_problem} is performed using the deterministic RMTR method, global convergence of which is shown in~\cite{Gratton2008recursive}. }

\begin{algorithm}
\footnotesize
\caption{\footnotesize DSS-RMTR($\L, \  \vec{\theta}_{0,0}^L, \ \Delta_{0,0}, \ \text{epoch}_{\max}, \text{mbs}_0$)}
\label{alg:DSSRMTR}
\begin{algorithmic}[1]
\Require{\footnotesize $\L:\R^{n} \rightarrow \R, \   {\vec{\theta}}_{0,0} \in \R^{n}, \ \Delta_0  \in \R,  \text{epoch}_{\max} \in \mathbb{N}, \text{mbs}_0 \in \N$}
\Constants {\footnotesize $o \in \R, \  L \in \N $}
\For{\footnotesize $e=0, ...,  \text{epoch}_{\max}$}
\State \footnotesize  $\{\D_b\}_{b=1}^{\text{n}_e}$ = GenMiniBatches($\D$, $\text{mbs}_e$,~$o$)  \Comment{\footnotesize Construct mini-batches (with overlap~$o$)}
\For{\footnotesize $\text{b}=1, ..., \text{n}_e$}
\State \footnotesize Generate $\L_b$  using $\D_{b}$   \Comment{\footnotesize Construct mini-batch objective function}
\State \footnotesize [$\vec{\theta}_{e, b}, \Delta_{e, b}$] = RMTR($L, \  \L_b, \  \vec{\theta}_{e, b-1}, \  \Delta_{e, b-1}$) \Comment{\footnotesize Call \cref{alg:rmtr_algo} }
\State \footnotesize $\text{red}_b = \L_b(\vec{\theta}_{e, b-1}) - \L_b(\vec{\theta}_{e, b})$ \Comment{ \footnotesize Compute mini-batch reduction}
\EndFor
\If{\footnotesize $\text{mbs}_e < |\D|$} \Comment{ \footnotesize Detect mini-batch (stochastic) regime }
\State \footnotesize  $\rho^G_e = \frac{\L(\vec{\theta}_{e,0}) - \L(\vec{\theta}_{e,\text{n}_e}) }{\frac{1}{\text{n}_e}  \sum_{b=1}^{\text{n}_e} \text{red}_b } $
\Comment{ \footnotesize Compute global (batch) TR ratio }
\State \footnotesize [$\vec{\theta}_{e+1, 0}, \text{mbs}_{e+1}$] = Gcontrol($\rho_e^G,  \ \vec{\theta}_{e, 0}, \  \vec{\theta}_{e, \text{n}_e}, \ \text{mbs}_e$)  \Comment{ \footnotesize Call \cref{alg:mb_conv_controll}}
\Else											\Comment{ \footnotesize Detect deterministic regime }
\State \footnotesize $\vec{\theta}_{e+1, 0}  = \vec{\theta}_{e, \text{n}_e}$	
\EndIf
\State \footnotesize  $\Delta_{e+1,0} = \Delta_{e,\text{n}_e}$  		\Comment{ \footnotesize Initialize TR radius for next epoch}
\EndFor
\State 
\Return \footnotesize $\vec{\theta}_{e+1,0}$, $\Delta_{e+1,0}$
\end{algorithmic}
\end{algorithm}

\textcolor{white}{.}\\
\begin{minipage}{0.48\linewidth}
\begin{algorithm}[H]
\footnotesize
\caption{\footnotesize ConvControl($\rho_i, \vec{\theta}_i,  \sv_i, \Delta_i$)}
\label{alg:conv_control}
\begin{algorithmic}[1]
\Require{\footnotesize $\rho_i \in \R, \    \vec{\theta}_i, \sv_i \in \R^{n},\ \Delta_i \in \R$}
\Constants {\footnotesize ${\Delta_{\text{min}}, \Delta_{\text{max}}}, \eta_1, \eta_2, \gamma_1, \gamma_2 \in \R$, \\
where $ 0< \eta_1 \leq \eta_2 < 1$ and $0 < \gamma_1 < 1 < \gamma_2$}
\If{\footnotesize $\rho_i > \eta_1$}        
\State \footnotesize $\vec{\theta}_{*} = \vec{\theta}_i + \sv_i $		\Comment{ \footnotesize Accept trial point}
\Else
\State \footnotesize  $\vec{\theta}_{*} = \vec{\theta}_i$				\Comment{\footnotesize Reject trial point}
\EndIf
\State \Comment{\footnotesize Adjust trust-region radius} 
\State $
\Delta_{*} =
\begin{cases}
{\max(\Delta_{\text{min}}}, \gamma_1  \Delta_{ i}), & \rho_i^l < \eta_1,  \\
\Delta_{i},  & \rho_i^l \in  [ \eta_1, \eta_2],   \\
{\min(\Delta_{\text{max}}}, \gamma_2 \Delta_{i}), & \rho_i^l >\eta_2, 
\end{cases}
$
\State \Return  \footnotesize	$\vec{\theta}_{*}, \Delta_{*}$
\end{algorithmic}
\end{algorithm}
\end{minipage}
\hfill
\begin{minipage}{0.49\linewidth}
\begin{algorithm}[H]
\footnotesize
\caption{\footnotesize Gcontrol($\rho_e^G, \vec{\theta}_{e}, \vec{\theta}_{e+1}, \text{mbs}_e$)}
\label{alg:mb_conv_controll}
\begin{algorithmic}[1]
\Require{\footnotesize $\rho_e^G \in \R, \   {\vec{\theta}}_e, \vec{\theta}_{e+1} \in \R^{n}, \ \text{mbs}_e \in \N$}
\Constants {\footnotesize $\zeta_1, \zeta_2, \omega \in \R$, where \\
$ \zeta_1>0, \  0<  \zeta_2 \leq 0.2, \  \omega > 1$}
\If{\footnotesize $\rho_e^G > \zeta_1$}        
\State \footnotesize $\vec{\theta}_{*} = \vec{\theta}_{e+1}$		\Comment{ \footnotesize Accept trial point}
\Else
\State \footnotesize  $\vec{\theta}_{*} = \vec{\theta}_{e}$				\Comment{\footnotesize Reject trial point}
\EndIf
\If{\footnotesize $\rho_e^G < \zeta_2$}					
\State \footnotesize $\text{mbs}_{*} = \omega \  \text{mbs}_{e} $      \Comment{ \footnotesize Increase mbs size}
\Else
\State \footnotesize $\text{mbs}_{*} =  \text{mbs}_{e}$			\Comment{ \footnotesize Preserve mbs size}
\EndIf 
\State \Return  \footnotesize	$\vec{\theta}_{*}, \text{mbs}_{*}$
\end{algorithmic}
\end{algorithm}
\end{minipage}

\paragraph{{Implementation}}
{The practical implementation of the DSS strategy using a single GPU requires a certain consideration, as the evaluation of loss/gradient for large mini-batches might be prohibitive due to the memory limitations.
We can overcome this difficulty by dividing the {large} mini-batch into smaller chunks of the data. 
The {large} mini-batch loss and gradient are then computed by aggregating the losses and gradients, evaluated using these smaller chunks.
Here, we highlight the fact that these chunks can be processed in parallel, for example using multiple GPUs. 
In this scenario, it is actually beneficial to use large mini-batches as soon as possible, in order to utilize all available resources and in turn to reduce the training time~\cite{shallue2018measuring}.}

\paragraph{{Hyper-parameter search}}
{The cost of standard training methods is traditionally very high, as one has to minimize~\eqref{eq:discrete_problem} multiple times, using different hyper-parameters, e.g., learning rate, and mini-batch size. 
The DSS-RMTR method proposed in this work overcomes this difficulty, as the step size is naturally induced by the trust-region radius. 
Secondly, the DSS strategy generates a sequence of appropriate mini-batch sizes during the training, depending on the observed progress.
We however note, that in order to achieve good generalization properties, {the} initial mini-batch size~$\text{mbs}_0$ should be sufficiently small. 
The simplest approach is to set $\text{mbs}_0$ to one and let the DSS strategy to adjust the mini-batch sizes appropriately within the first few epochs. }

\subsection{DSS-RMTR method with limited-memory quasi-Newton Hessian approximation} 
\label{sec:mini_batches_genration}
The convergence speed of the DSS-RMTR method can be enhanced by incorporating the curvature information. 
In this work, we approximate a Hessian on all levels of the multilevel hierarchy using the L-SR1 method.
Given a level $l$, the L-SR1 method considers a memory ${\{\sv_i^l, \zv_i^l\}_{i=1}^M}$ of~$M$ secant pairs.
Each secant pair~$\{\sv_i^l, \zv_i^l\}$ consists of a search direction~$\sv_i^l$ and the variation of the gradient along this direction, denoted by~$\zv_i^l$.
Typically, the secant pairs~$\{\sv_i^l, \zv_i^l\}_{i=1}^M$ are collected during the iteration process over last~$M$ iterations. 
In the context of our DSS-RMTR method, this would mean that the pair~$\{\sv_i^l, \zv_i^l\}$ is obtained as
\begin{equation}
\begin{aligned}
\sv_i^l &= \vec{\theta}_{b, i+1}^l - \vec{\theta}_{b, i}^l, \\
\zv_i^l &= \nabla \L_b^l (\vec{\theta}_{b, i+1}^l) - \nabla \L_b^l (\vec{\theta}_{b, i}^l),
\end{aligned}
\label{eq:up_ne_qn_1}
\end{equation}
where~$\sv^l_i$ is a search-direction computed at level~$l$, during the~$i$-th {iteration}, while minimizing a local objective function~$\L_b$.
The vector~$\zv^l_i$ expresses the difference between the gradients of the local objective function~$\L_b^l$, evaluated at~$\vec{\theta}^l_{b, i+1}$ and~$\vec{\theta}^l_{b, i}$.

Unfortunately, evaluating~$\zv^l_i$ as in~\eqref{eq:up_ne_qn_1} immensely increases the computational cost of our multilevel method. 
For example, let us assume that the RMTR method is set up with one pre- and one post-smoothing step on a level~$l$. 
One V-cycle then requires two gradient evaluations per smoothing step, i.e., four gradient evaluations per level.
In contrast, usage of the first-order smoother would require only two gradient evaluations per level.

\subsubsection{Generating mini-batches with overlap}
\label{sec:overlapping_minibatches}
\begin{figure}[t]
	\centering	
\scalebox{0.8}{		
\includegraphics{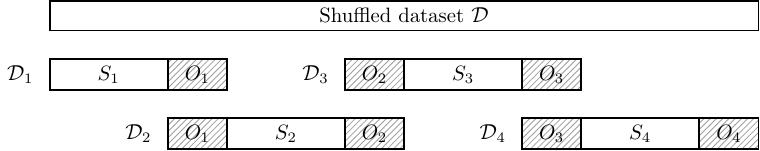}	
    }
    \caption{Example of four mini-batches created with overlap.}
    \label{fig:subdomains_of_batches}
\end{figure}

We can decrease the computational cost associated with the evaluation of the secant pairs~${\{\sv_i^l, \zv_i^l\}_{i=1}^M}$ by utilizing an overlapping sampling strategy.
This strategy was originally proposed to ensure the stability of limited-memory quasi-Newton updates in stochastic settings~\cite{berahas2020robust, erway2020trust}. 
The main idea behind this method is to split a shuffled dataset~$\D$ into~$n_e$ mini-batches of size~$\text{mbs}_e$. 
Each mini-batch~$\D_b$ is constructed as~$\D_b=\{ O_{b-1}, S_b, O_b \}$, where~$S_b$ denotes samples unique to the mini-batch~$\D_b$. 
Symbols~$O_{b-1}, O_b$ denote samples of mini-batch~$\D_b$, which are shared with mini-batches~$\D_{b-1}$ and~$\D_{b+1}$, respectively. 
The number of overlapping samples contained in~$O_{b-1}, O_b$ is usually fairly low.
In this work, we prescribe~$20\%$ overlap between samples in~$\D_b$ and~$\D_{b+1}$, for all~$b \in \{1, \ldots, n_e-1\}$ during the first epoch.
This determines the size of~$O_{b-1}$, and $O_b$, which we then keep constant during the whole training. 
Thus, the ratio between an overlapping and a non-overlapping portion of the samples in mini-batch increases during training. 
\Cref{fig:subdomains_of_batches} illustrates the construction of mini-batches using the overlapping sampling strategy.

Now, we can  evaluate $\zv_i^l$ on a given level~$l$ as follows:
\begin{equation}
\begin{aligned}
\zv_i^l &= \nabla \L_{O_b}^l (\vec{\theta}_{b, i+1}^l) -  \nabla \L_{O_b}^l (\vec{\theta}_{b, i}^l),
\end{aligned}
\label{eq:up_ne_qn}
\end{equation}
where~$\nabla \L_{O_b}^l$ denotes a gradient of~\eqref{eq:discrete_problem}, evaluated using only samples contained in~$O_b$. 
Given that ${|O_b| < |\D_b|}$, the evaluation of~$\zv_i^l$ using~\eqref{eq:up_ne_qn} is computationally cheaper than using formula~\eqref{eq:up_ne_qn_1}. 
In addition, the gradients~$\nabla \L_{O_b}^L (\vec{\theta}_{b, i+1}^L) $ evaluated during the post-smoothing step on the finest level can be utilized to compute~${\nabla \L_{\D_{b+1}}^L (\vec{\theta}_{b+1, 0}^L)}$ during the pre-smoothing step of the next V-cycle.

\section{Numerical experiments}
\label{sec:numerical_experiments}
In this section, we describe numerical examples, which we use to assess the convergence properties of the proposed DSS-RMTR method. 
Our numerical examples consider regression and classification tasks. 
The regression tasks are associated with surrogate modeling of the PDEs, while the classification tasks deal with image recognition. 

All presented numerical examples are associated with solving the optimization problem~\eqref{eq:discrete_problem}.
The multilevel hierarchy of ResNets, required by the RMTR method,  is obtained by performing the time discretization with varying discretization parameters. 
Below, we provide a description of the network architecture associated with the coarsest level, where~$l=1$.
Deeper networks are obtained by uniform refinement with a factor of two, as described in \cref{sec:ml_hierarchy_resnet}.
{Description of the ResNets associated with different levels of the multilevel hierarchy for all numerical examples can be found in \cref{table:levels_description}.}

\begin{table}
  \centering
          \footnotesize    
          \caption{{The number of ResNet parameters associated with different levels of the multilevel hierarchy. }}
  \label{table:levels_description}
  \begin{tabular}{ c|cccccccccccc}

\multirow{2}{*}{{{Example}}}         &    \multicolumn{6}{c}{{{Level}}}    \\ 
                      				&   {1} 				& {2}     		& {3} 				& {4}  		& {5} & {6}  \\ \hline \hline 
{NDR}  					&    	{$1,395$}		& {$2,355$}    		& {$4,275$} 			&{ $8,115$}  	& {$15,795$} &{ $31,155$} \\				
{TDD}  					&    	{$1,290$}		& {$2,250$}    		& {$4,170$} 			& {$8,010$}  	& {$15,690$} & {$31,050$} \\				

{Fashion}  					&    	{$466,714$}	& {$855,578$}    & {$1,633,306$} 		& {$3,188,762$} & $--$ & $--$ \\
{CIFAR-10{/CIFAR-100}}  				&    	{$1,082,426$}	& {$1,860,154$}    & {$3,415,610$} 	& {$6,511,226$}	& $-$ & $--$  
                                                   
\end{tabular}
\end{table}

\subsection{Regression tasks {- dense networks}}
\label{sec:reg_tasks}
Many engineering systems are modeled by partial differential equations (PDEs), which are parametrized by a large number of design/input parameters. 
To evaluate such models, a large number of simulations have to be performed for various designs.
As a consequence, tasks such as sensitivity analysis, or uncertainty quantification, become computationally infeasible as they necessitate solution of a large number of numerical simulations.  
Surrogate models alleviate this computational burden by constructing approximate models, which are significantly cheaper to evaluate. 
Here, we employ a data-driven approach and construct surrogates using ResNets. 
The networks are trained to approximate the response of the simulator for given input parameters. 
Training is performed using a dataset of parameter-observable pairs and the least-squares loss, defined as
$\ell(\yv_s, \cv_s) = \| \yv_s - \cv_s \|^2_2,$
where~$\yv_s$ is a prediction made by the ResNet, and $\cv_s$ is an observable for given input parameters~$\xv_s $.

We investigate two examples, datasets of which were generated by solving the underlying PDEs using the finite element (FE) framework MOOSE~\cite{gaston2009moose}. 
During our experiments, we consider ResNets with $5$ residual blocks, $T=5$ and $\beta_1=\beta_2=10^{-4}$ on the coarsest level. 
Each residual block has the form of a single layer perceptron, i.e.,~${\pazocal{F}(\vec{\theta}_k, \vec{q}_k): = \sigma( \Wm_k \vec{q}_k + \bv_k)}$, where ${\vec{\theta}_k=(\text{flat}(\Wm_k),  \text{flat}(\bv_k))}$, with $\Wm_k \in \R^{5 \times 5}$, and $\bv_k \in \R^{5}$. 
The activation function~$\sigma$ is chosen as \emph{tanh}.

\subsubsection{Time-dependent diffusion (TDD)}
This example considers the time-dependent diffusion equation defined on the time interval $[0, 1]$ and spatial domain $\Omega = (-0.5, 0.5)^2$, with boundary $\Gamma = [-0.5, 0.5]^2 \setminus \Omega$. 
The formulation of the problem is given as
\begin{equation}
\begin{aligned}
\frac{\partial \psi}{\partial t} - \nabla \cdot \bigg( D \bigg[  \frac{300}{\psi}\bigg]   \nabla \psi    \bigg) &= 1,000 \sin(f t) \ \mathbbm{1}_{\Omega_S}(\xv),  \qquad &  \text{on} \  \Omega \times (0, 1], \\ 
- D \bigg[  \frac{300}{\psi}\bigg]^2 \nabla \psi \cdot \nv &= 0, & \text{on} \  \Gamma \times (0, 1], \\ 
\psi &= \psi_0, \quad & \text{on} \  \Omega \times \{ 0 \} ,\\ 
\end{aligned}
\label{eq:tdd}
\end{equation}
where $\psi: \Omega \times [0, 1] \rightarrow \R $ is a state variable expressing the temperature and the symbol $\xv$ denotes spatial coordinates. 
The indicator function $\mathbbm{1}_{\Omega_S}(\xv): \R^2 \rightarrow \{0, 1\}$ takes on value $1$ if $\xv \in \Omega_S$ and $0$ otherwise. 
Here, $\Omega_S$ indicates the source region defined inside of the domain $\Omega$ as $\Omega_S = [-0.1, 0.1]^2$. 

Equation~\eqref{eq:tdd} is parametrized by the initial temperature~${\psi_0}$, the frequency multiplier~${f}$ and the diffusion coefficient~${D}$. 
We are interested in obtaining a surrogate, which is capable of predicting maximum and minimum temperatures over both spatial and temporal domains. 
Thus, given input features $\xv_s = [\psi_0, f, {D}]$, the ResNet is trained to predict ${\cv_s=[\psi_{\text{max}}, \psi_{\text{min}}]}$, where
$\psi_{\text{max}} = \max_{\xv \in \Omega, t \in [0,1]} \psi(\xv, t)$ and $\psi_{\text{min}} = \min_{\xv \in \Omega, t \in [0,1]} \psi(\xv, t)$.
We generate a dataset consisting of $2,000$ samples, $1,600$ for training, and $400$ for validation, by repeatedly solving the PDE numerically. 
In particular, we discretize~\eqref{eq:tdd} in space using the FE method on a quadrilateral mesh with $200$ nodes in each spatial dimension. 
The time discretization is performed using the explicit Euler method with $100$ time-steps. 
The parameters~$\psi_0, f,  {D}$ are sampled from a uniform distribution, see \cref{tab:params_surrogate} for details. 
An example of simulation results for different values of $\psi_0, f, {D}$ is illustrated in \cref{fig:surrogate_examples} on the right.

\subsubsection{Neutron diffusion-reaction (NDR)}
\begin{figure}
\includegraphics[scale=0.21]{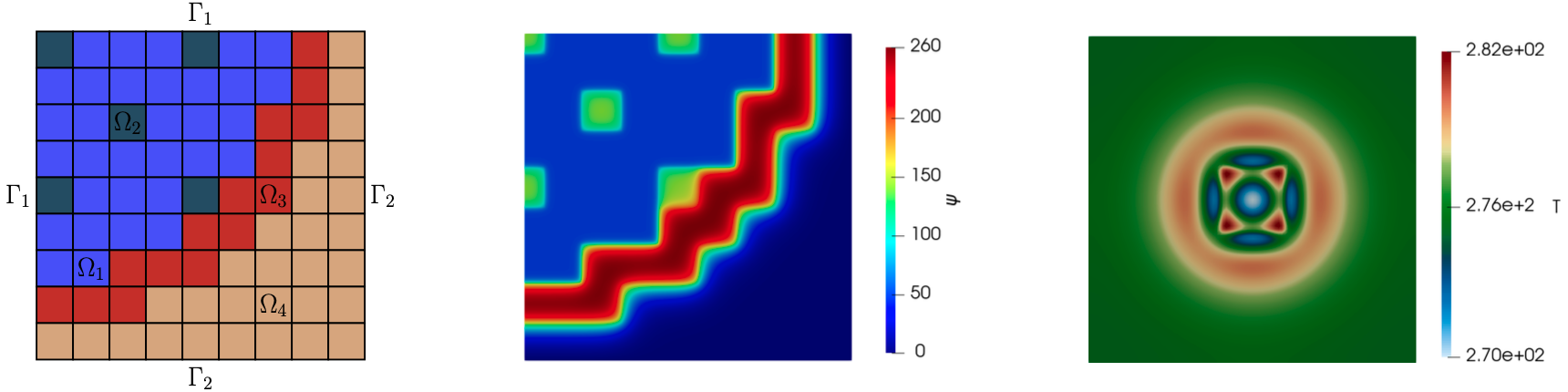}
  \caption{
  \emph{Left:} The geometry used for the NDR example. Domain $\Omega$ is decomposed into four subdomains, illustrated by different colors. 
  \emph{Middle/Right:}    An example of simulation results used for the generation of the NDR and the TDD dataset, respectively.}
    \label{fig:surrogate_examples}
  \end{figure}

Following~\cite{prince2019parametric}, we consider a steady-state neutron diffusion-reaction problem with spatially-varying coefficients and an external source.
As common for nuclear reactor modeling, the domain $\Omega=(0, 170)^2$ is heterogeneous and consists of four different material regions, denoted by $\Omega_1, \ldots, \Omega_4 $, and depicted on \cref{fig:surrogate_examples} on the left.  
The strong-form of the problem is given as 
\begin{align}
 \nabla \cdot [ D(\xv) \nabla \psi(\xv) ] +  \alpha (\xv) \nabla \psi(\xv)  &= q(\xv), \qquad &\text{on} \ \Omega, \nonumber \\ 
\psi(\xv) &= 0, &\text{on} \ \Gamma_{1} := [0, 170] \times \{1\} \cup  \{0\} \times [0, 170] , \label{eq:neutron}\\ 
 D(\xv) \nabla \psi(\xv) \cdot \nv(\xv) &= 0 &\text{on} \ \Gamma_{2} := [0, 170] \times \{0\} \cup  \{1\} \times [0, 170] , \nonumber
\end{align}
where~$\psi:\Omega \rightarrow \R$ is the neutron flux (scalar quantity) and $\xv$ denotes spatial coordinates.
Functions~$D, \alpha, q$ are defined as ${D(\xv) = \sum_{i=1}^4 \mathbbm{1}_{\Omega_i}(\xv) D_i}$, ${q(\xv) = \sum_{i=1}^3 \mathbbm{1}_{\Omega_i}(\xv) q_i}$, and ${\alpha(\xv) = \sum_{i=1}^4 \mathbbm{1}_{\Omega_i}(\xv) \alpha_i}$.
Here, the indicator function $\mathbbm{1}_{\Omega_i}(\xv): \R^2 \rightarrow \{0, 1\}$ takes on value $1$, if $\xv \in \Omega_i$ and $0$ otherwise.  
Problem~\eqref{eq:neutron} is parametrized by the 11 parameters, i.e., diffusion coefficients~$\{ D_i \}_{i=1}^4$, reaction coefficients~$\{ \alpha_i \}_{i=1}^4$ and sources~$\{ q_i \}_{i=1}^3$. 

We aim to construct a surrogate that can predict the average neutron flux over the whole domain $\Omega$. 
Thus, given input parameters $\xv_s \in \R^{11}$, the network is trained to approximate $\cv_s = \hat{\psi}$, where $\hat{\psi} = \frac{\int_{\Omega} \psi(\xv) \ d\xv }{\int_{\Omega} d\xv } $. 
We generate the dataset of $3, 000$ samples, which we split to $2,600$ for training and $400$ for testing.
The details regarding the distributions of sampled parameters can be found in \cref{tab:params_surrogate}. 
The resulting PDEs are solved using the FE method on a quadrilateral mesh, which consists of $500$ nodes in both spatial dimensions.

\begin{table}
\caption{Distributions of the parameters used for generation of TDD and NDR datasets.
A uniform distribution~$\pazocal{U}(a,b)$ is specified by lower (a) and upper (b) bounds. }
\label{tab:params_surrogate}
\centering
\small
\begin{tabular}{cl|cl|cl}

\multicolumn{2}{c|}{TDD} & \multicolumn{4}{c}{NDR} \\ \hline
 Param. & Distribution  &  Param. & Distribution &  Param. & Distribution  \\ \hline
$D$ & $\sim \pazocal{U}(0.01, 0.02)$ & $\{D_i\}_{i=1}^3$ & $\sim \pazocal{U}(0.15, 0.6)$ & $\alpha_1$ & $\sim \pazocal{U}(0.0425, 0.17)$    \\ \hline
$f$ & $\sim \pazocal{U}(15, 25)$   & $D_4$ & $\sim \pazocal{U}(0.2, 0.8)$  & $\alpha_2$ & $\sim \pazocal{U}(0.065, 0.26)$    \\ \hline
$\psi_0$ & $\sim \pazocal{U}(270, 330)$  & $\{q_i\}_{i=1}^3$ & $\sim \pazocal{U}(5, 20)$    & $\alpha_3$ & $\sim \pazocal{U}(0.04, 0.16)$    \\ \hline
& & $q_4$ & $\sim \pazocal{U}(0, 0)$  & $\alpha_4$ & $\sim \pazocal{U}(0.005, 0.02)$    \\  
\end{tabular}
\end{table}

\subsection{Classification tasks - {convolutional networks}}
Furthermore, we consider classification tasks, using convolutional ResNets.
Training is performed using the softmax hypothesis function and the cross-entropy loss function, defined as
${\ell(\yv_s, \cv_s) = \cv_s^T \log (\yv_s)}$, where~$\yv_s$ denotes class-label probabilities returned by the ResNet and~$\cv_s$ denotes the target given by the dataset. 

{Here, we use residual blocks, which have} {the following form: ${\pazocal{F}(\vec{\theta}_k, \vec{q}_k): = \sigma( \text{BN}( \Wm_{k_1}  ( \sigma( \text{BN}( \Wm_{k_2} \vec{q}_k))))}$, where~$\sigma$ denotes  \emph{ReLu} activation function and $\text{BN}$ stands for the batch normalization~\cite{ioffe2015batch}.}
In contrast to previous sections, the matrices~{$\mat{W}_{k_1}, \mat{W}_{k_2}$, for all ${k=0, \ldots, K-1}$} now represent sparse convolutional operators.

We consider the following datasets of images:
\begin{itemize}
\item \textbf{Fashion:}
Fashion-MNIST dataset contains images of cloth pieces grouped into~$10$ classes~\cite{xiao2017fashion}. 
The dataset consists of~$60,000$ grayscale images for training and~$10,000$ for validation. 
We perform the classification using a three-stage ResNet, {recall \cref{sec:nets_with_var_width}}.
At the beginning of each stage, we double the number of filters and halve the size of the feature map.
The dimensionality of the feature map is preserved for all layers within a given stage. 
We employ the same time discretization parameters for all stages.
Thus, the depth of the coarse-level network is defined by~$T_i=3$ and~{$K_i=3$}, where~$i \in \{1, 2, 3 \}$. 
The number of filters is chosen to be~$16$,~$32$, and~$64$.
The operators~$\{\Qm_i\}_{i=1}^{3}$, which provide a change of dimensionality between different stages, represent an average pooling operation~\cite{goodfellow2016deep}. 
The regularization parameters are chosen as~$\beta_1=6 \times 10^{-4}$ and $\beta_2=10^{-4}$.
\item \textbf{CIFAR-10 { and CIFAR-100}:} 
{The CIFAR-10 and CIFAR-100 datasets consist of~$60,000$ color images, where~$50,000$ is designated for training and~$10,000$ for validation~\cite{krizhevsky2009learning} .}
 {Each~$32\times32$ image belongs to one of~$10$ and $100$ classes for CIFAR-10 and CIFAR-100, respectively.}
We employ the same three-stage ResNet architecture as for the Fashion dataset{, but the number of filters is chosen to be~$32$,~$64$, and~$128$.  Moreover, the regularization parameters are set to~$\beta_1= 5 \times 10^{-4}$ and $\beta_2=10^{-3}$.}
\end{itemize}
{All three} datasets are pre-processed by standardizing the images,  so that pixel values lie in the range~$[0, 1]$ {and by subtracting the mean from each pixel.}
{In addition, we make use of standard data augmentation techniques, i.e., image rotation, horizontal and vertical shift and horizontal flip. }

 \section{Numerical results}
\label{sec:num_results}
In this section, we study the convergence properties of the proposed DSS-RMTR method. 
Our implementation of ResNets is based on the library Keras~\cite{chollet2015keras}, while the solution strategies are implemented using library NumPy~\cite{walt2011numpy}. During all experiments, we consider a fixed set of parameters, summarized in \cref{table:params_solver_resnet}.
The choice of parameters~$\eta_1, \eta_2, \gamma_1, \gamma_2$ follows common practice in the trust-region literature, see for instance~\cite{Conn2000trust}. 
The parameters~$\zeta_1, \zeta_2, \omega$ are selected in accordance with~\cite{mohr2019adaptive}. 
The parametric and algorithmic choices specific to the RMTR method reflect our numerical experience, acquired using a model problem, see the supplement (\cref{sec:suplement_RMTR}). 
More precisely, the RMTR method is configured as F-cycle with one pre/post-smoothing step and three coarse-level steps.
{The minimum and maximum radii $\Delta_{\text{min}}$ and $\Delta_{\text{max}}$ are set to constant values, namely $10^{-7}$ and $0.5$, for all numerical examples except CIFAR-100. 
For CIFAR-100 dataset, we decrease $\Delta_{\text{max}}$ by a factor of $5$ every time the new level is incorporated into the multilevel hierarchy within the F-cycle. 
Although this slightly increases the computational cost of the proposed RMTR method, it gives rise to models with higher validation accuracy. }
At the end, we also highlight the fact that we incorporate the momentum term into our trust-region multilevel framework, see { \cref{sec:momentum_ml} for details.}

The single-level DSS-TR method is obtained by calling the DSS-RMTR algorithm with~$L=1$. 
The numerical experiments employ the DSS-RMTR method with and without the Hessian approximation strategy. 
If only first-order information is used, then the solution of the trust-region subproblem is provided by a Cauchy point (CP)~\cite{nocedal2006numerical}. 
If the LSR1 Hessian approximation is employed, then the trust-region subproblems are solved using the orthonormal basis method~\cite{brust2017solving}.
Our implementation of the LSR1 method is based on a compact matrix representation~\cite{nocedal1980updating}.
An initial approximation of the Hessian is obtained by solving an eigenvalue problem as proposed in~\cite{rafati2018improving}.

Compared to the first-order stochastic methods, the limited-memory secant methods have a higher computational cost per iteration. 
However, this additional cost becomes marginal as the size of mini-batches increases~\cite{bottou2018optimization}. 
For this reason, we set the memory size to~$M=1$ at the beginning of the training process. 
The value of~$M$ is increased by one, every time the mini-batch size is enlarged by the DSS strategy.  

\begin{table}
  \centering
          \small  
  \caption{Choice of parameters used inside TR/RMTR algorithms.}
  \label{table:params_solver_resnet}
  \begin{tabular}{ c|cccc|ccc|ccc|ccccc}
    {Parameter} & {$\eta_1$} & {$\eta_2$} & {$\gamma_1$} & {$\gamma_2$} & $\zeta_1$ & $\zeta_2$ & $\omega$ & {$\mu_s$}  & {$\mu_s$} &  ${\vartheta}$ & ${\Delta_{\text{min}}}$ & ${\Delta_{\text{max}}}$  \\  \hline 
    {Value}     & $0.1$      & $0.75$     & $0.5$        & $2.0$        & $0.1$     & $0.0$     & $2.0$   & $1$       &  {$3$}  & ${0.9}$ & {$10^{-7}$} & {0.5}  \\ 
\end{tabular}
\end{table}

All presented experiments are performed at the Swiss National Supercomputing Centre~(CSCS) using XC50 compute nodes of the Piz Daint supercomputer.
Each XC50 compute node consists of the Intel Xeon E5-2690 v3 processor and an NVIDIA Tesla P100 graphics card.
The memory of a node is $64\,$GB, while the memory of a graphics card is $16\,$GB.

To assess the performance of the methods, we provide a comparison with the single-level (DSS-)TR method {and with two baseline methods, namely SGD and Adam, implementation of which is provided by the Keras framework.
The hyper-parameters for both baseline methods have been found by hyper-parameter search reported in the supplement (\cref{sec:hyper_tunning_adam_gd}).}

Since the computational cost of one cycle of the RMTR method is higher than the computational cost of one TR/SGD/Adam iteration, we need to devise a suitable metric to perform a fair comparison.
We focus on the most expensive part of the training, i.e., the cost associated with an evaluation of the gradients. 
To this aim, we define one work unit~$W^L$ to represent a computational cost associated with an evaluation of the gradient on the finest level, using a full dataset~$\D$.
Given that the computational cost of the back-propagation algorithm scales linearly with the number of samples and the number of the layers, we can define the total computational cost~$W$ as follows:
\begin{align}
 W = \sum_{e=1}^{e_{\text{tot}}} \sum_{b=1}^{n_e} \sum_{l=1}^L  \frac{n_b}{p}  2^{l-L} Q^l_b W^L,
 \label{eq:W_rmtr_mini_batch}
\end{align}
where~$e_{\text{tot}}$ denotes the number of epochs required for convergence. 
{The symbol~$Q^l_b$ describes a number of gradient evaluations performed on a given level~$l$ for the mini-batch~$\D_b$.}
Given an epoch~$e$, the computational cost is obtained by summing up gradient evaluations performed on all levels using all mini-batches. 
Since the computational cost of a gradient evaluation on level~$l <L$ using mini-batch~$\D_b$ is lower than one work unit~$W^L$, we need to rescale quantities in~\eqref{eq:W_rmtr_mini_batch} accordingly. 
In particular, the scaling factor~$2^{l-L}$ accounts for the difference between the computational cost on a level~$l$ and the finest level~$L$. 
Please note that this scaling factor assumes a uniform coarsening in 1D by a factor of two. 
The scaling factor ~$ \frac{n_b}{p}$ takes into consideration the difference between the number of samples contained in the dataset~$\D$ and the mini-batch~$\D_b$.

\subsection{Regression tasks}
\label{sec:reg_results}
In this section, we study the convergence properties of {all training methods} using regression tasks associated with surrogate modeling of PDEs. 
These type{s} of problems are often solved using large batches, or even {the} full dataset. 
As a consequence, we investigate the performance of the  method{s} only in deterministic settings. 
{All} solution strategies terminate, if the following stopping criterion: 
$W > W_{\max}$,  is satisfied. 
Here, the symbol~$W_{\max}$ denotes a fixed budget for which we can afford to train the networks. 
The value of $W_{\max}$ is prescribed to $600$ and $1,000$ for the TDD and the NRD example, respectively. 

\Cref{tab:surrogate_results} depicts the obtained results in terms of training and validation loss achieved after the training.
The results are gathered for ResNets with $129$ residual blocks (6 levels). 
{Note that it is quite common to employ shallower, but wider networks for solving such regression tasks. 
Although ResNets considered here are thinner and deeper, they contain approximately the same number of trainable parameters. 
In addition, their structure allows us to create a multilevel hierarchy and study the convergence properties of the proposed RMTR method. 
}

\begin{figure}[t]
\centering
\includegraphics{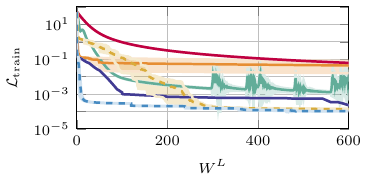}
\includegraphics{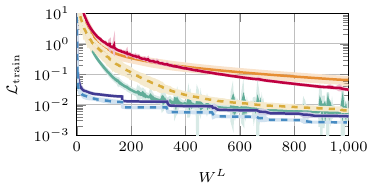}
\includegraphics{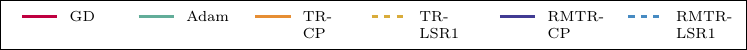}
\caption{{ Mean training loss and $95\%$ confidence interval over 10 independent runs.
Experiments performed using TDD/NRD dataset (\emph{Left/Right}).}}
\label{fig:surrogate_plots}
\end{figure}

As we can see, employing the Hessian approximation strategy is beneficial for both single-level TR and RMTR methods. 
For instance, the TR method provides approximately two orders of magnitude more accurate solution if the LSR1 method is used. 
{We can also observe that the Adam method outperforms the standard GD method for both examples, but achieves lower train and test loss than TR-CP for the TDD example.  }
{
The numerical results presented in \cref{fig:surrogate_plots} also demonstrate that the RMTR method always provides a solution with a lower value of the loss function.
The obtained difference is especially prevalent at the beginning of the solution process. 
As expected, the improvement factor obtained by the RMTR method is larger if the LSR1 is employed. 
However, the RMTR variant without Hessian approximation performs significantly better than other first-order methods, i.e.,~TR-CP and GD method. 
In the end, we also point out that the standard deviation of the obtained results is lower if the multilevel method is used, compared to Adam and single-level TR methods. 
Hence, the performance of the RMTR method is more stable with respect to the choice of initial parameters. 
}

\begin{table}
\centering
\small
\caption{Mean training and validation loss $\pm$ standard deviation for regression tasks.
Results obtained over $10$ independent runs. 
{Experiments performed with a prescribed computational budget $W_{\text{max}}=600\  W^L$ and $W_{\text{max}}=1, 000 \ W^L$ for the TDD and the NDR example, respectively.} }
\label{tab:surrogate_results}
  \begin{tabular}{ r|cc|ccccccccc}
\multicolumn{1}{c|}{\multirow{2}{*}{Method}}	  &  \multicolumn{2}{c|}{TDD}  &  \multicolumn{2}{c}{NDR}  \\ 
& $ \L_{\text{train}} (\times 10^{-4})$     & $\L_{\text{test}}  (\times 10^{-4})$ &    	$\L_{\text{train}}$    &   $\L_{\text{test}}$    \\ \hline \hline
{GD}	& 	${608.6 \pm 0.06}$  &	${610.4 \pm 0.37 }$ 			& 	${0.0314 \pm 0.002}$  &	${0.0319 \pm 0.003 }$ \\ \hline
{Adam} & 		${42.66 \pm 8.36}$  &	${49.82 \pm 13.44}$		& 	${0.0054 \pm 0.127}$  &	${0.0087 \pm 0.271}$  \\ \hline

{TR-CP}	& 	${425.9 \pm 121.3}$  &	${452.7 \pm 151.8 }$ 	& 	{$0.0655 \pm 0.043$}  &	{$0.0795 \pm 0.107$} \\
{TR-LSR1} & 		${1.33 \pm 0.42}$  &	{$1.41 \pm 0.31$}		& 	{$0.0065\pm 0.007$}  &	{$0.0072 \pm 0.018$}  \\ \hline

{RMTR-CP }  	& 		{$2.17 \pm	0.05$} & 	{$2.32	\pm 0.15$}	& 	{$0.0041 \pm	0.002$} & {	$0.0045	\pm 0.008$}  \\
{RMTR-LSR1} 	& 		${ 1.02 \pm	0.01}$ & 	${1.21 \pm 0.03}$		& 	${0.0026 \pm	0.001}$ & 	${0.0031 \pm	0.006}$  \\ \end{tabular}
\end{table}

\subsection{Classification - convolutional networks}
\label{sec:results_classification}
Our {next} set of experiments tackles image-reconginition with convolutional neural networks. 
Here, we consider only hybrid (stochastic-deterministic) settings, as the convergence of the deterministic methods is very poor for these examples. 
During all experiments, we prescribe an initial mini-batch size~$\text{mbs}_0$ to be~{$100$ and $256$ for CIFAR-10{/CIFAR-100} and Fashion dataset, respectively}. 
Since convolutional ResNets are more challenging to train than dense ResNets, we employ {the} following stopping criterion:
\begin{align*}
 &\bigg(\sum_{i=1}^{10}  (\text{acc}_{\text{train}})_e  -  (\text{acc}_{\text{train}})_{e-i}  \bigg) < 0.001 \  && \text{or}  \  \bigg(\sum_{i=1}^{10}  (\text{acc}_{\text{val}})_e  -  (\text{acc}_{\text{val}})_{e-i} \bigg)  < 0.001  \\
 & \qquad \qquad \qquad \qquad \quad \qquad \text{acc}_{\text{train}} > 0.99 \quad && \text{or} \qquad \qquad \qquad \qquad \qquad \ \  \text{acc}_{\text{val}} > 0.99. \label{eq:acc_stoping_conv}
\end{align*}
{
The train accuracy $\text{acc}_{\text{train}}$ and validation accuracy $\text{acc}_{\text{val}}$ are defined as
\begin{align*}
\text{acc}_{\text{train/val}} = \frac{\text{number of correctly classified samples from the train/val. dataset}}{\text{total number of samples in the train/val. dataset}}.
\end{align*}
}
This stopping criterion verifies whether a training or validation accuracy of~$99\%$ is achieved. 
In addition, it incorporates early stopping, which halts the training process when there is no improvement in training or validation accuracy within the last~$10$ epochs~\cite{goodfellow2016deep}.

We investigate the convergence properties of the {training} methods by measuring the total computational cost and achieved training and validation accuracy. 
{Tables~\ref{tab:stochastic_test_convolutional_nets_new_fashion}, \ref{tab:stochastic_test_convolutional_nets_new_cifar10} and \ref{tab:stochastic_test_convolutional_nets_new_cifar100} summarize the obtained results for the Fashion, CIFAR-10 and CIFAR-100 datsets with respect to increasing number of residual blocks.}
{For all three datasets, we train all networks~$10$ times and report the best result in terms of validation accuracy $\text{acc}_{\text{val}}$, i.e., the results which generalize the best.
In order to gain insight into the sensitivity of the obtained results, we also provide the average $\text{acc}_{\text{val}}$ and $W$, together with their standard deviations.}

The obtained results indicate that ResNets with more residual blocks {and corresponding more levels} can achieve higher validation accuracy, in turn justifying the higher computational cost. 
We can also observe that the SGD method is capable of achieving higher validation accuracy than the Adam method, for {all three} datasets. 
This is in agreement with the numerical experience reported in the literature, see for example~\cite{keskar2017improving, loshchilov2017decoupled}. 
We also note that for the Fashion dataset, the trust-region methods with the LSR1 Hessian approximation strategy are capable of achieving slightly higher accuracy, compared to their first-order counterparts. 
For the CIFAR-10 {and CIFAR-100} datasets, we detect the opposite behavior, i.e., the DSS-TR-LSR1 and DSS-RMTR-LSR1 methods are more prone to overtraining and therefore achieve lower validation accuracy than DSS-TR-CP and DSS-RMTR-CP, respectively. 
{This is contrary to the results obtained for the regression tasks in \cref{sec:reg_results}, where the use of LSR1 Hessian approximation helped to achieve the results with higher accuracy, i.e., lower value of the loss.}

The obtained results also demonstrate that the SGD method requires the highest computational cost amongst all solution strategies. 
Interestingly, this difference is more prevalent for the Fashion example. 
For the CIFAR-10 example, SGD, Adam, and DSS-TR-CP require a comparable computational cost to satisfy the prescribed convergence criteria. 
Interestingly, employing the LSR1 approximation strategy within trust-region methods is more beneficial for the Fashion example. 
For instance, the DSS-TR-LSR1 method is approximately $4$ times faster than DSS-TR-CP. 
An opposite trend is observed for the CIFAR-10 dataset, as the use of the LSR1 approximation strategy causes an increase in the computational cost of the DSS-RMTR method. 
In this particular case, the use of the LSR1 approximation scheme does not improve the convergence of the DSS-RMTR method.
However, it comes at a higher computational cost, as several gradient computations are required for the evaluation of secant pairs, recall \cref{sec:overlapping_minibatches}. 
Thus, we can infer that employing Hessian approximation strategies for classification tasks is not as beneficial as for more ill-conditioned regression tasks considered in~\cref{sec:reg_results}.

The results reported in Tables~\ref{tab:stochastic_test_convolutional_nets_new_fashion}, \ref{tab:stochastic_test_convolutional_nets_new_cifar10} and \ref{tab:stochastic_test_convolutional_nets_new_cifar100} also demonstrate that the DSS-RMTR method outperforms all other training strategies, in terms of computational cost. 
Moreover, the obtained speedup grows with the number of levels. 
For instance, DSS-RMTR-LSR1 achieves speedup by a factor of $5.6$ compared to SGD, for two levels and the Fashion dataset. 
At the same time, the speedup by a factor of $12.9$ can be observed for four levels. 
Similarly, for the CIFAR-10 dataset, the DSS-RMTR-CP method achieves speedup compared to the SGD method by a factor of $1.9$ for two levels, but by a factor of $3.6$ for four levels.

\begin{figure}[t]
\centering
\includegraphics{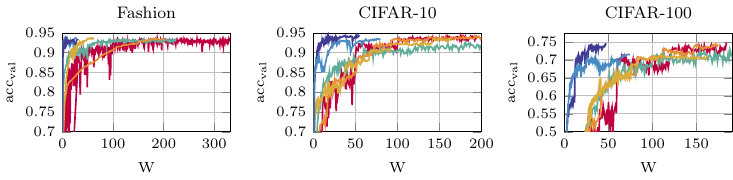}
\includegraphics{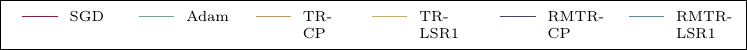}
\caption{{The validation accuracy for convolutional ResNets, associated with four levels. 
The best result in terms of validation accuracy was selected from $10$ independent runs.}}
  \label{fig:conv_results}
\end{figure}

{The obtained results also demonstrate that the convergence of the DSS-RMTR methods is less sensitive to the choice of initial guess. 
In particular, the standard deviation of validation accuracy $\text{acc}_{\text{val}}$ as well as the computational cost is significantly lower for DSS-RMTR methods compared to single-level methods. }
Moreover, we can also observe that the DSS-RMTR method approaches a higher accuracy much more quickly, due to the good initial guess obtained from the coarser levels, {see also \cref{fig:conv_results}.}
This is of particular interest for the training scenarios with a fixed computational budget. 
In the end, we also point out that the training using trust-region methods is not subjected to hyperparameter tuning, thus reducing their overall computational cost immensely, in comparison with the SGD and the Adam methods.

\begin{table}
  \centering
          \footnotesize    
          \caption{{The validation accuracy $\text{acc}_{\text{val}}$ and total computational cost $W$ of the solution strategies required for training convolutional ResNets for Fashion dataset.
The best and average validation accuracy $\text{acc}_{\text{val}}$ obtained from $10$ independent runs. }}
  \label{tab:stochastic_test_convolutional_nets_new_fashion}
  \begin{tabular}{ c|c|cc|cccccc}
{Method}                    &     {L}     &  {$\text{best acc}_{\text{val}}$}  & {{W}}  & {$\text{avg. acc}_{\text{val}}$}  & {{avg. W}}       \\ \hline \hline

    \multirow{3}{*}{{{SGD} }} 	& {$2$}        &     	{$93.33\%$} 	& {$309$}  &     {$(93.12\pm 0.24)\%$}  &   {$312 \pm 21.3$} \\
                               		&{$3$}    	&     	{$93.53\%$}  	& {$319$}  	&   {$(93.38\pm 0.19)\%$}  &	  {$315 \pm 32.9$} \\
                               		& {$4$}     	& 	{$93.71\%$}  	& {$332$}	 & {$(93.61\pm 0.15)\%$} & {$341 \pm 18.1$}	 \\ \hline                                                             

    \multirow{3}{*}{{{Adam} }} & {$2$}       &        {$93.31\%$}      	& {$182$}          &  {$(92.98 \pm 0.19)\%$} &  {$179 \pm 18.3$} \\
                               		& {$3$}     	&       	{$93.40\%$} 		& {$228$} &  {$(93.29 \pm 0.14)\%$} & {$211 \pm 18.6$} \\		
                               		& {$4$}     	&        	{$93.50\%$} &         {$224$}  & {$(93.41 \pm 0.12 )\%$} & {$230 \pm  23.7$} \\ \hline

    \multirow{3}{*}{{{DSS-TR-CP} }} & {$2$}          & {$92.96\%$} & {$172$}        	&  {$(92.87 \pm 0.08)\%$} &  {$175 \pm 7.4$} \\
                               		& {$3$}     	   & {$93.12\%$} & {$184$}  	 			&  {$(93.11 \pm 0.06)\%$} &  {$179 \pm 8.9$} \\
                               		&{$4$}    	   & {$93.41\%$} & {$197$}   					&  {$(93.39 \pm 0.03)\%$} &  {$201 \pm 8.3$} \\ \hline

    \multirow{3}{*}{{{DSS-TR-LSR1} }} 	& {$2$}   		&    	{$93.32\%$} & {59}	&  {$(93.18 \pm 0.13)\%$} &  {$55 \pm 10.2$} \\
                               			& {$3$}     	&   {$93.43\%$ }	&{$63$}  &  {$(93.39 \pm 0.11)\%$} &  {$58 \pm 9.8$} \\
                               			& {$4$}     	& 	{$93.70\%$} 	& {$56$}    &  {$(93.58 \pm 0.07)\%$} &  {$54 \pm 9.1$} \\ \hline

    \multirow{3}{*}{{{DSS-RMTR-CP} }} 
    			& {$2$} 	& { $93.09\%$} 	& {$64.8$} 		&  {$(93.08 \pm 0.04)\%$} &  {$61 \pm 3.4$} \\
                         & {$3$}  	& {$93.23\%$}   	& {$48.3$} 		&  {$(93.22 \pm 0.02)\%$} &  {$47 \pm 4.1$} \\
                         & {$4$}    	& {$93.41\%$}  	&{$33.2$}  		&  {$(93.38 \pm 0.02)\%$} &  {$33 \pm 2.9$} \\ \hline

    \multirow{3}{*}{{{DSS-RMTR-LSR1} }} 	
    					& {$2$}     & {$93.12\%$}  	& {$55.0$}  		&  {$(93.07 \pm 0.08)\%$} &  {$52 \pm 3.9$} \\
                               		& {$3$}     & {$93.47\%$}      	& {$45.9$} 		&  {$(93.41 \pm 0.06)\%$} &  {$47 \pm 3.3$} \\
                               		& {$4$}     &  {$93.69\%$} 	&  {${25.7}$ }  		&  {$(93.62 \pm 0.09)\%$} &  {$25 \pm 1.9$} \\

  \end{tabular}
\end{table}

\begin{table}
  \centering
          \footnotesize    
          \caption{{The validation accuracy $\text{acc}_{\text{val}}$ and total computational cost $W$ of the solution strategies required for training convolutional ResNets for CIFAR-10 dataset.
The best and average validation accuracy $\text{acc}_{\text{val}}$ obtained from $10$ independent runs.}}
  \label{tab:stochastic_test_convolutional_nets_new_cifar10}
  \begin{tabular}{ c|c|cc|cccccc}
{Method}                    &     {L}     &  {$\text{best acc}_{\text{val}}$}  & {{W}}  & {$\text{avg. acc}_{\text{val}}$}  & {{avg. W}}       \\ \hline \hline

    \multirow{3}{*}{{{SGD} }} 	& {$2$}        &     	{$93.98\%$} 	& {$201$}  &     {$(93.31\pm 0.72)\%$}  &   {$189 \pm 42.5$} \\
                               		&{$3$}    	&     	{$94.06\%$}  	& {$184$}  	&   {$(93.74\pm 0.48)\%$}  &	  {$192 \pm 37.3$} \\
                               		& {$4$}     	& 	{$94.32\%$}  	& {$195$}	 & {$(93.93\pm 0.39)\%$} & {$201 \pm 34.8$}	 \\ \hline                                                             

    \multirow{3}{*}{{{Adam} }} & {$2$}       &        {$92.52\%$}      	& {$164$}          &  {$(92.32 \pm 0.19)\%$} &  {$158 \pm 16.2$} \\
                               		& {$3$}     	&       	{$92.55\%$} 		& {$177$} &  {$(92.38 \pm 0.16)\%$} & {$174 \pm 16.1$} \\		
                               		& {$4$}     	&        	{$92.62\%$} &         {$181$}  & {$(92.55 \pm 0.12)\%$} & {$193 \pm  16.5$} \\ \hline                                              
               
    \multirow{3}{*}{{{DSS-TR-CP} }} & {$2$}          & {$93.67\%$} & {$189$}        	&  {$(93.72 \pm 0.07)\%$} &  {$191 \pm 6.5$} \\
                               		& {$3$}     	   & {$93.94\%$} & {$193$}  	 			&  {$(93.88 \pm 0.09)\%$} &  {$190 \pm 7.1$} \\
                               		&{$4$}    	   & {$94.18\%$} & {$203$}   					&  {$(94.13 \pm 0.07)\%$} &  {$195 \pm 6.9$} \\ \hline

    \multirow{3}{*}{{{DSS-TR-LSR1} }} 	
    			& {$2$}   		&    	{$92.91\%$} 	& {157}		&  {$(92.80 \pm 0.12)\%$} &  {$161 \pm 15.3$} \\
                         & {$3$}     	&   	{$93.28\%$ }	&{$165$}  	&  {$(93.23 \pm 0.09)\%$} &  {$160 \pm 13.4$} \\
                         & {$4$}     	& 	{$93.45\%$} 	& {$162$}    	&  {$(93.37 \pm 0.08)\%$} &  {$154 \pm 11.7$} \\ \hline

    \multirow{3}{*}{{{DSS-RMTR-CP} }} 
    			& {$2$} 	& { $93.72\%$} 	& {$102.3$} 		&  {$(93.71 \pm 0.03)\%$} &  {$101 \pm 3.2$} \\
                         & {$3$}  	& {$94.11\%$}   	& {$71.7$} 		&  {$(94.09 \pm 0.03)\%$} &  {$71 \pm 2.6$} \\
                         & {$4$}    	& {$94.37\%$}  	&{$54.8$}  		&  {$(93.32 \pm 0.06)\%$} &  {$49 \pm 5.7$} \\ \hline

    \multirow{3}{*}{{{DSS-RMTR-LSR1} }} 	
    					& {$2$}     & {$93.03\%$}  	& {$140.4$}  		&  {$(93.01 \pm 0.11)\%$} &  {$143 \pm 5.1$} \\
                               		& {$3$}     & {$93.28\%$}      	& {$97.2$} 		&  {$(93.21 \pm 0.09)\%$} &  {$95 \pm 3.9$} \\
                               		& {$4$}     &  {$93.55\%$} 	&  {${79.9}$ }  		&  {$(93.48 \pm 0.10)\%$} &  {$77 \pm 3.6$} \\

  \end{tabular}
\end{table}

\begin{table}
  \centering
          \footnotesize    
          \caption{{The validation accuracy $\text{acc}_{\text{val}}$ and total computational cost $W$ of the solution strategies required for training convolutional ResNets for CIFAR-100 dataset.
The best and average validation accuracy $\text{acc}_{\text{val}}$ obtained from $10$ independent runs.}}
  \label{tab:stochastic_test_convolutional_nets_new_cifar100}
  \begin{tabular}{ c|c|cc|cccccc}
\multicolumn{1}{c|}{Method}                   &     {L}     &  {$\text{best acc}_{\text{val}}$}  & {{W}}  & {$\text{avg. acc}_{\text{val}}$}  & {{avg. W}}       \\ \hline \hline

    \multirow{3}{*}{{{SGD} }} 	& {$2$}        &     	{$72.10\%$} 	& {$147$}  &     {$(71.51\pm 0.44)\%$}  &   {$198 \pm 55.7$} \\
                               		&{$3$}    	&     	{$73.63\%$}  	& {$228$}  	&   {$(73.25\pm 0.35)\%$}  &	  {$231 \pm 25.2$} \\
                               		& {$4$}     	& 	{$74.74\%$}  	& {$184$}	 & {$(73.79\pm 0.69)\%$} & {$211 \pm 56.4$}	 \\ \hline                                                             

    \multirow{3}{*}{{{Adam} }} & {$2$}       &        {$69.81\%$}      	& {$180$}          &  {$(69.18 \pm 1.24)\%$} &  {$181 \pm 12.3$} \\
                               		& {$3$}     	&       	{$71.42\%$} 		& {$165$} &  {$(69.83 \pm 0.46)\%$} & {$177 \pm 16.7$} \\		
                               		& {$4$}     	&        	{$72.96\%$} &         {$188$}  & {$(69.02 \pm 1.43)\%$} & {$187 \pm  10.5$} \\ \hline

    \multirow{3}{*}{{{DSS-TR-CP} }} & {$2$}          & {$72.31\%$} & {$141$}        	&  {$(69.24 \pm 2.56)\%$} &  {$166 \pm 25.6$} \\
                               		& {$3$}     	   & {$73.68\%$} & {$158$}  	 			&  {$(71.38 \pm 2.41)\%$} &  {$174 \pm 17.4$} \\
                               		&{$4$}    	   & {$74.59\%$} & {$175$}   					&  {$(73.82 \pm 0.70)\%$} &  {$183 \pm 14.1$} \\ \hline

    \multirow{3}{*}{{{DSS-TR-LSR1} }} 	
    			& {$2$}   		&    	{$69.91\%$} 	& {156}		&  {$(68.59 \pm 1.82)\%$} &  {$163 \pm 31.2$} \\
                         & {$3$}     	&   	{$70.73\%$ }	&{$192$}  	&  {$(69.37 \pm 2.64)\%$} &  {$207 \pm 15.7$} \\
                         & {$4$}     	& 	{$71.14\%$} 	& {$162$}    	&  {$(70.30 \pm 1.33)\%$} &  {$188 \pm 28.5$} \\ \hline

    \multirow{3}{*}{{{DSS-RMTR-CP} }} 
    			& {$2$} 	& { $72.04\%$} 	& {$97.4$} 		&  {$(71.48 \pm 0.21)\%$} &  {$101 \pm 5.6$} \\
                         & {$3$}  	& {$73.71\%$}   	& {$58.8$} 		&  {$(73.59 \pm 0.13)\%$} &  {$68 \pm 3.7$} \\
                         & {$4$}    	& {$74.56\%$}  	&{$47.9$}  		&  {$(74.52 \pm 0.09)\%$} &  {$42 \pm 2.1$} \\ \hline

    \multirow{3}{*}{{{DSS-RMTR-LSR1} }} 	
    					& {$2$}     & {$69.95\%$}  	& {$138.1$}  		&  {$(69.71 \pm 0.33)\%$} &  {$157 \pm 7.7$} \\
                               		& {$3$}     & {$71.04\%$}      	& {$99.9$} 		&  {$(70.89 \pm 0.28)\%$} &  {$101 \pm 8.1$} \\
                               		& {$4$}     &  {$71.53\%$} 	&  {$74.7$}  		&  {$(71.35 \pm 0.24)\%$} &  {$83 \pm 8.8$} \\

  \end{tabular}
\end{table}

\subsection{{Execution time}}
{
All aforementioned numerical results are reported in terms of the computational cost $W$. 
Using this measure allows us to investigate the asymptotic behavior of the training algorithms without considering their implementation aspects.
This is of particular importance, as the SGD and Adam methods have benefited from years of code optimization by a large user and developer base of the Keras library.
In contrast, the trust-region based methods considered in this work were developed by the authors with the sole purpose to conceptualize and assess the proposed RMTR method.
}

{
To demonstrate the performance of our implementation of the RMTR method, we also report the execution time for one particular numerical example, namely training of three-stage convolutional ResNet with $9$ residual blocks per stage (3 levels) for the CIFAR-10 dataset. 
As we can see from Table~\ref{tab:timing}, the RMTR-CP method achieves an approximate speedup by a factor of $2$ compared to SGD and Adam methods. 
Although, this result roughly corresponds to the work unit estimate, the execution time per one work unit is higher for the trust-region based methods.
This is caused not only by the lack of code optimization but also due to the fact that the trust-region algorithms are algorithmically more elaborate than SGD/Adam methods. 
More precisely, they require an (approximate) solution to the trust-region sub-problem and an evaluation of the trust-region ratio $\rho$, which in turn increases the computational cost.
However, these additional computations allow for the construction of globally convergent methods, which do not require a user-prescribed learning rate.
Instead, the learning rate is induced automatically by means of the trust-region radius. 
In contrast, the learning rate of SGD and Adam methods is typically found during the hyper-parameter search, which significantly increases the reported execution time (in our case by a factor of $16$).}

\begin{table}
\centering
          \captionof{table}{{The execution time required to train ResNet with $9$ residual blocks per stage for the CIFAR-10 dataset. 
          The example considers the multilevel hierarchy with three levels. The time is reported in seconds.}}
  \label{tab:timing}
  \begin{tabular}{ c|c|c|ccccccc}
       \multirow{2}{*}{ {Method} }				&  {Work units } 		& {Total time} 	& {Time per work unit}  \\ 
 											&  {(W)} 			& { (T)} 			& { (T/W)}  \\ \hline \hline
{SGD} & 				{184.0} 		& {143.52}			& {0.78}		\\
{Adam} &  				{177.0} 		& {155.76}			& {0.88}		\\ \hline
{DSS-TR-CP} &  			{193.0} 		& {187.21}			& {0.97}		\\   
{DSS-TR-LSR1} &  		{165.0} 		& {206.25}			& {1.25}		\\  \hline
{DSS-RMTR-CP} &  		{71.7}		& {72.42}			& {1.01}		\\  
{DSS-RMTR-LSR1} &  		{97.2}		& {130.25}			& {1.34}		\\  
  \end{tabular}
\end{table}

\section{Conclusion}
\label{sec:conclusion}
In this work, we proposed a novel variant of the RMTR method, specifically tailored for training ResNets. 
Our multilevel framework utilized a hierarchy of auxiliary networks with different depths to speed up the training process of the original network. 
The proposed RMTR method operated in a hybrid (stochastic-deterministic) regime and dynamically adjusted mini-batch sizes during the training process. 
Furthermore, we incorporated curvature information on each level of the multilevel hierarchy using the limited-memory SR1 method. 
The numerical performance of the proposed multilevel training method was presented on regression and classification tasks. 
A comparison with a SGD, Adam and single-level TR method was performed and illustrated a significant reduction in terms of the computational cost. 
We also demonstrated that the RMTR method is considerably less sensitive to the choice of the initial guess and typically produces a more accurate solution, for a fixed computational budget.

The presented work can be extended in several ways. 
{For instance, it would be beneficial} 
to incorporate adaptive time refinement techniques and the integrator refinement strategies. 
For the convolutional neural networks, we also aim to explore a coarsening in space (image resolution). 

\appendix
\section{{Incorporating momentum into TR/RMTR framework}}
\label{sec:momentum_ml}
{
Let $\vv_i^l \in \R^{n^l}$ be the momentum term, defined as 
$\vv_i^l = \vartheta \vv_{i-1}^l +\sv_i^l$,
where~${\vartheta \in \R}$. 
Following~\cite{erway2020trust1}, $\vv_i^l$ is crafted into the trust-region framework by modifying the search direction $\sv_i^l$ obtained by solving the trust-region subproblem~\eqref{eq:tr_subproblem} as follows}
\begin{equation}
\begin{aligned}
{\sv_i^l = \min \bigg(1.0, \frac{\Delta_i^l}{\| \vv_i^l  + \sv_i^l \|} \bigg)(\vv_i^l + \sv_i^l ),}
\end{aligned}
\end{equation}
{where $\vv_i^l = \vartheta \min \bigg(1.0, \frac{\Delta_i^l}{\|  \vv_{i-1}^l \|} \bigg) \vv_{i-1}^l $.}

{For multilevel settings, we keep track of the momentum by transferring it across the multilevel hierarchy.
{In this way, the search directions computed on every level take into account the history of updates, which has been accumulated over all levels.}
More precisely, we initialize $\vv_0^{l-1}$ during the coarse-level parameter initialization phase as $\vv_0^{l-1} = \bold{\Pi}_l^{l-1} \vv_{\mu_s}^{l}$. 
After the coarse-level solve is performed, the updated coarse-level momentum $\vv_{\mu^l}^{l-1}$ is then used to update $ \vv_{\mu_s}^{l}$ as $ \vv_{\mu_s +1}^{l} = \vv_{\mu_s}^{l} + \Pm_{l-1}^l (\vv_{\mu^{l-1}}^{l-1} - \vv_0^{l-1})$. 
This step takes place at the same time as the prolongation of coarse-level correction. 
A similar approach for transferring the momentum across the multilevel hierarchy was considered in the context of the full approximation scheme in~\cite{ponce_ML2022}. 
}

\section{{Incorporating batch normalization and data augmentation into TR/RMTR framework}}
\label{sec:data augmentation}
{The batch normalization and data augmentation break the finite-sum structure of the loss function. 
As a consequence, the trust-region methods, which rely on the monotonic decrease of the objective function, cannot be readily applied.
Here, we describe an alternative approach, suitable for multilevel trust-region framework.}

\subsection{{Data augmentation}}
{The standard data augmentation approach is to form a class of transform functions. 
On each iteration, a particular transform is randomly selected and the gradient is evaluated for transformed data. 
The transforms are typically applied as a part of the data pipeline.
However, using different transforms for each evaluation of the loss function or gradient prohibits the convergence control provided by the trust-region algorithms.  
We can ensure that the method provides a local monotonic decrease in loss function by 
selecting one particular transform at the beginning of each V-cycle, for a given mini-batch.
This transform is then used for all subsequent evaluations of the loss and the gradient within the V-cycle. }

\subsection{{Batch normalization}}
{Batch normalization (BN) layers normalize the output of the activation function, denoted by $x$. 
This is achieved by applying the following transformation:
\begin{align}
y = \frac{x - {E}[x]}{\sqrt{\text{Var}[x]+ \epsilon}}\gamma + \beta,
\label{eq:bn_eq}
\end{align}
where $\gamma, \beta$ are learnable parameters and $E[\cdot], \text{Var}[\cdot]$ denote the expectation and the variance, calculated over a given mini-batch. 
Since test data might not be mini-batched or might originate from different distribution as training data, one also has to keep track of exponential moving mean and variance (MMV). 
At inference, MMV is used instead of mini-batch statistics.
In the context of TR methods, the evaluation of loss and gradient for one mini-batch occurs at multiple points. 
This causes MMV to be updated using statistics obtained at multiple points, resulting in a loss of convergence. 
As a remedy, we evaluate mini-batch statistics and update  MMV only during the first loss/gradient evaluation, for a given mini-batch. 
During all other evaluations, we reuse the precomputed mini-batch statistics,  prohibit updating MMV, but allow parameters $\gamma, \beta$ to be updated. }

{In the multilevel settings, we update MMV only at the beginning of each V-cycle, i.e.,~only during the first finest level evaluation of the loss/gradient. 
BN layers on all levels are switched to inference mode. 
Thus, the training of $\gamma, \beta$ is allowed only on the finest level. 
On all other levels, mini-batch statistics and parameters $\gamma, \beta$ are obtained by projecting the quantities from the finest level.  
In this way, we ensure that the coarse-level corrections are consistent with the finest level. 
Furthermore, we point out that an additional synchronization of the mini-batch statistics is required for an evaluation of global $\rho^G_e$. }

\section{Detailed numerical investigation of the properties of the RMTR method}
\label{sec:suplement_RMTR}
\begin{figure}
\centering
\includegraphics[scale=0.3]{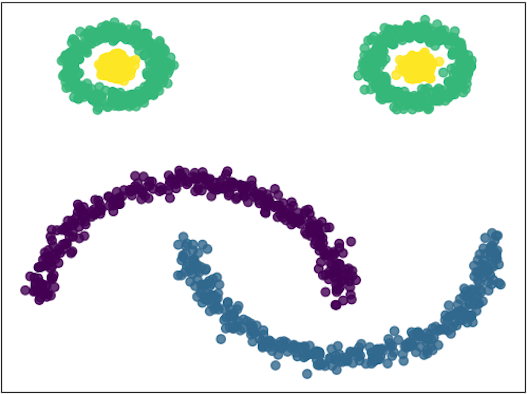}
\hspace{1.5cm}
\includegraphics[scale=0.3]{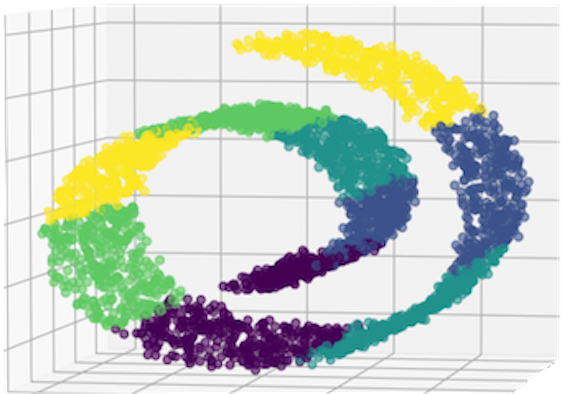}
\caption{\emph{Left/Right:} Smiley/Spiral dataset consisting of 4/5 classes (illustrated by different colors). }
\label{fig:simple_datasets_supplement}
\end{figure}

{In this section, we study the convergence properties of the (DSS-)RMTR method with respect to the algorithmic choices.
More precisely, the focus is given to the choice of projection operator, cycling strategy, number of smoothing/coarse-level steps, use of momentum, and the choice of initial mini-batch size.}
To this aim, we consider ResNets with dense single-layer perceptron residual blocks, and two artificially created datasets, which contain particles located in 2D/3D.
Thus, the input features describe the coordinates of the particle, while the output vector prescribes an affiliation to a given class. 
Although these datasets do not capture real-life applications, they allow us to investigate the convergence properties of the proposed DSS-RMTR method at a low computational cost. 
In particular, we employ the following datasets:
\begin{itemize}
\item \textbf{Smiley:}
The smiley dataset contains particles of the two-dimensional plane~$[-5, 5]^2$ categorized into~$4$ classes. 
Each class is related to a particular part of Smiley, see~\cref{fig:simple_datasets_supplement} on the left. 
The dataset consists of~$7,000$ samples, divided into~$5,000$ for training and~$2,000$ for validation.
We use a network architecture with a \emph{tanh} activation function, fixed-width of~$10$, and a depth of~$K=7$ on the coarsest level. 
The value of~$T$ is prescribed as~$T=1$ and $\beta_1=\beta_2= 10^{-4}$.
\item \textbf{Spiral:}
The spiral dataset incorporates particles in a three-dimensional hyperplane~$[-1.5, 1.5]^3$, classified to~$5$ classes. 
All particles are located on spiral geometry, which is generated as described in~\cite[Chapter 10]{marsland2015machine} and implemented within the Sklearn library~\cite{pedregosa2011scikit}.
The position of particles on a spiral defines~$10$ unique chunks.  
Each chunk is assigned randomly to one of~$5$ classes, such that each class consists of two unique chunks. 
The spiral dataset contains~$7,000$ samples, where~$5,000$ are used for training and~$2,000$ are used for validation purposes. 
During this experiment, we use  a ResNet with a fixed width of~$5$ and a \emph{tanh} activation function. 
The network depth is defined by $K=7$, and $T=7$, while regularization parameters are chosen as~$\beta_1=\beta_2= 5 \times 10^{-4}$.
\end{itemize}

\subsection{Numerical results}
\label{sec:supl_rmtr_params}
{
The presented investigation of the convergence properties of the (DSS-)RMTR method is divided into two parts, related to deterministic and stochastic settings.}
During all experiments, we employ the following stopping criterion: 
\begin{align*}
  \text{acc}_{\text{train}} > 0.98 \quad \text{or} \quad \text{acc}_{\text{val}} > 0.98, 
\end{align*}
where~$\text{acc}_{\text{train}}$ and~$\text{acc}_{\text{val}}$ denote the training and the validation accuracy, respectively.
{The reported experiments also include the comparison of the (DSS-)RMTR method with its single-level counterpart, the (DSS-)TR method.}

\subsubsection{Deterministic settings}
Our first set of experiments is associated with deterministic settings and the performance of the TR/RMTR method with respect to the increasing number of residual blocks/levels. 
During these experiments, both TR and RMTR methods employ LSR1 Hessian approximation, implemented in conjunction with overlapping sampling strategy presented in~\cref{sec:overlapping_minibatches}.

\paragraph{Projection operators}
{The appropriate choice of the projection operator ${\bold{\Pi}}_{l-1}^l$ is crucial for the efficiency of the RMTR method~\cite{Gross2009, gross2009unifying}.
This is due to the fact that a quantity which gets transferred by the RMTR method to the finer level is a coarse-level correction, defined as}
\begin{align*}
{\vec{s}^{l} = \vec{\theta}^{l}_* - \vec{\theta}_0^{l},}
\end{align*}
{where $\vec{\theta}^{l}_*$ and $\vec{\theta}_0^{l}$ denote the obtained solution and the initial guess on level $l$, respectively. 
Note, that  by the definition, $\vec{s}^{l}$ depends on the initial guess~$\vec{\theta}_0^{l} := \vec{\Pi}^{l}_{l+1} \vec{\theta}_{\mu_1}^{l+1}$, obtained by means of the operator~$\vec{\Pi}^{l}_{l+1} $.
Thus, the use of different projection operators leads to different coarse-level corrections and, therefore, to different fine-level trial points.
As a consequence, employing the projection operator with poor approximation properties might slow down the overall convergence of the multilevel method. }

{Here, we investigate three possibilities:}
\begin{enumerate}
\item {$\vec{\Pi}_{l}^{l-1} =  \big( ( \vec{P}_{l-1}^{l})^T \vec{P}_{l-1}^{l} \big)^{-1} ( \vec{P}_{l-1}^{l})^T $ (Moore--Penrose pseudo-inverse of  $ \vec{P}_{l-1}^{l}$).}
\item { $\vec{\Pi}_{l}^{l-1} = ( \vec{P}_{l-1}^{l})^T$ (adjoint of prolongation operator $ \vec{P}_{l-1}^{l}$).}
\item {$\vec{\Pi}_{l}^{l-1} = \vec{D} ( \vec{P}_{l-1}^{l})^T$ (scaled adjoint of prolongation operator $ \vec{P}_{l-1}^{l}$).}
\end{enumerate} 

{The first option represents an "ideal" choice, as it is designed to satisfy the following requirement:
\begin{align}
\vec{\theta}^{l-1}  &= \vec{\Pi}^{l-1}_{l} \underbrace{(\vec{P}_{l-1}^{l} \vec{\theta}^{l-1} )}_{\vec{\theta}^{l}},
\label{eq:identity_operation}
\end{align}
which states that transferring the parameters to the subsequent level of multilevel hierarchy and back does not result in their alteration.
An operator~$\vec{\Pi}^{l}_{l+1}$ that meets requirement~\eqref{eq:identity_operation} can be found by solving the following least-square minimization problem:
\begin{equation}
\vec{\theta}^{l-1} := \underset{\vec{\theta}^{l-1}}{\text{min}} \ \| \vec{\theta}^{l} - \vec{P}_{l-1}^{l} \vec{\theta}^{l-1} \|^2,
\label{eq:least_square}
\end{equation}
which gives rise to
$\vec{\theta}^{l-1} = ((\vec{P}_{l-1}^{l} )^T \vec{P}_{l-1}^{l} )^{-1} (\vec{P}_{l-1}^{l})^T \vec{\theta}^{l}$.
Hence, the operator~$\vec{\Pi}^{l-1}_{l}$ is uniquely obtained as
\begin{align}
\vec{\Pi}^{l-1}_{l} = ((\vec{P}_{l-1}^{l} )^T \vec{P}_{l-1}^{l} )^{-1} (\vec{P}_{l-1}^{l})^T,
\label{eq:P_ls}
\end{align}
i.e.,~as the Moore--Penrose pseudo-inverse of~$\vec{P}_{l-1}^{l} $. }

{Unfortunately, in order to evaluate~\eqref{eq:P_ls}, the solution of a linear system is required, which makes the multilevel algorithm computationally expensive.  
Moreover, the resulting operator~$\vec{\Pi}^{l}_{l+1}$ is a dense matrix, even though the matrix~$\vec{P}_{l-1}^{l}$ is sparse. 
As a consequence, it is desirable to employ computationally cheaper alternatives. 
In the literature, it is quite common to use the adjoint of prolongation operator, i.e., $(\vec{P}_{l-1}^{l})^T$. 
This is however not ideal, as $(\vec{P}_{l-1}^{l})^T$ is a dual operator~\cite{hackbusch2013multi}, designed to transfer the dual quantities, such as gradients. 
In the context of ResNets, this particular choice of the transfer operator causes an increase in the magnitude of the weights, which can in turn lead to numerical instabilities, such as exploding gradients. 
More suitable alternative is to employ the appropriately scaled $(\vec{P}_{l-1}^{l})^T$, i.e., $\vec{D} ( \vec{P}_{l-1}^{l})^T$, where $\vec{D}$ is a diagonal matrix. 
It has been demonstrated in~\cite{kopanivcakova2020multilevel}, that if $\vec{D}$ is chosen well, then $\vec{D} ( \vec{P}_{l-1}^{l})^T$ closely approximates $((\vec{P}_{l-1}^{l} )^T \vec{P}_{l-1}^{l} )^{-1} (\vec{P}_{l-1}^{l})^T$. 
For instance, if $\vec{P}^{l}_{l+1}$ is constructed as standard interpolation, assembled for uniform meshes in 1D, scaling $( \vec{P}_{l-1}^{l})^T$ with the uniform factor $0.5$ is appropriate.  }

{
To demonstrate the computational cost of the RMTR method with respect to different projection operators, we perform a set of numerical experiments using the Spiral example. 
As we can observe from~\cref{table:deterministic_test_transfer_operators}, the use of Moore--Penrose pseudo-inverse gives rise to the most efficient RMTR variant. 
The highest computational cost is required when $\bold{\Pi}_{l-1}^{l}: = (\Pm_{l-1}^l)^T$ is employed. 
This is not surprising, as this particular choice of $\bold{\Pi}_{l-1}^{l}$ causes an increase in the magnitude of the weights on the coarser levels. 
Here, we would like to highlight the fact that even though the computational cost, i.e., the number of gradient evaluations increases, the RMTR method remains globally convergent.
This is due to the fact, that the coarse-level corrections which increase the fine-level loss are discarded by the algorithm, recall~\cref{sec:rmtr_algo}.
We can also see, that the RMTR method configured with $\Dm(\Pm_{l-1}^l)^T$ yields comparable performance as the variant with the Moore--Penrose pseudo-inverse.
Since $\Dm(\Pm_{l-1}^l)^T$ requires lower computational cost and memory resources, we employ $\bold{\Pi}_{l-1}^{l}:= \Dm(\Pm_{l-1}^l)^T$ in order to generate all other numerical results presented in this work.}

\begin{table}
  \centering
    \small
  \caption{ {The average total computational cost ($W^L$) required by the deterministic RMTR method, configured as a V-cycle with the LSR1 Hessian approximation strategy, for the Spiral dataset.
The RMTR method is set up using three different types of the projection operator ${\bold{\Pi}}_{l-1}^l$.
 The results are obtained by averaging~$10$ independent runs.  } }
  \label{table:deterministic_test_transfer_operators}
  \begin{tabular}{ c|llllllll}
 \multicolumn{1}{c|}{{  \multirow{2}{*}{{$\bold{\Pi}_{l-1}^{l}$}}}}          &     \multicolumn{4}{c}{{Levels (Residual blocks)}}     \\         
&\multicolumn{1}{c}{{3 (25)}}  & \multicolumn{1}{c}{{4 (49)}}                 & \multicolumn{1}{c}{{5 (97)} }                 & \multicolumn{1}{c}{{6 (193)}}                                         \\ \hline \hline
{$(\Pm_{l-1}^l)^T$} & $  \textcolor{white}{0}{45.7 \pm} {7.9}$ & $\textcolor{white}{0}{55.9 \pm} {5.3}$ & $ \textcolor{white}{0}{98.8 \pm} {12.0}$ & ${156.4  \pm} {12.7}$ \\ 
{$\Dm(\Pm_{l-1}^l)^T$} & $  \textcolor{white}{0}{35.5 \pm} {4.7}$ & $\textcolor{white}{0}{43.2 \pm} {5.1}$ & $ \textcolor{white}{0}{82.4 \pm} {11.4}$ & ${131.4  \pm} {13.9}$ \\
{$\big((\Pm_{l-1}^l)^T (\Pm_{l-1}^l) \big)^{-1} (\Pm_{l-1}^l)^T$} & $  \textcolor{white}{0}{33.1 \pm} {4.1}$ & $\textcolor{white}{0}{39.9 \pm} {5.3}$ & $ \textcolor{white}{0}{83.1 \pm} {12.5}$ & ${126.8  \pm} {11.4}$ \\ 
  \end{tabular}
\end{table}

\begin{figure}
  \centering
\includegraphics{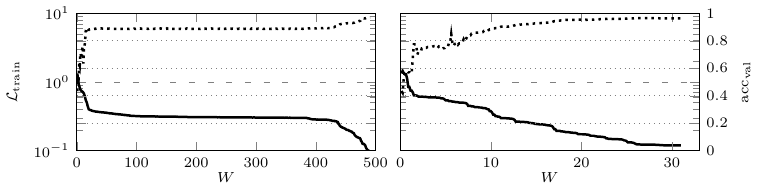}  
  \caption
  {Typical convergence behavior of the TR and the RMTR-F method when used for training of dense ResNets.
  The example considers a network with~$25$ residual blocks and the Smiley dataset.
  Validation accuracy is depicted by dotted lines, while training loss is depicted by solid lines. 
    \emph{Left:} The TR method.
    \emph{Right:} The four-level RMTR method. 
  }
  \label{fig:behavior_smiley}
\end{figure}

\paragraph{Cycling scheme}
{As a next step, we investigate the performance of the RMTR method with respect to the choice of cycling scheme. 
}
\cref{table:deterministic_test} reports the obtained results in terms of the average total computational cost and the standard deviation obtained over~$10$ independent runs. 
As we can see, the total computational cost of the TR method grows rapidly with the network depth. 
This behavior is expected, since it is known that deep networks are more difficult to train than shallow networks~\cite{haber2017stable}. 
\cref{fig:behavior_smiley} on the left depicts the typical convergence behavior of the TR method, used for the training of ResNets. 
We observe that the method encounters a certain plateau region, where only a small decrease in the value of the loss is obtained. 

Results reported in \cref{table:deterministic_test} also demonstrate how the choice of the cycling scheme influences the performance of the RMTR method. 
As we can see, the F-cycle is computationally less expensive than the V-cycle.
Besides, using F-cycle helps to reduce the variability of the obtained results. 
Therefore, for the remainder of this work, we use the RMTR method in the form of F-cycle.

In contrast to the TR method, the computational cost of the RMTR method in form of F-cycle decreases with the number of layers. 
This is due to the fact that the initialization of the network parameters, provided by the F-cycle, produces an initial guess which is relatively close to a solution. 
The plateau regions are typically encountered on the coarser levels, where the computational cost is low. 
The typical convergence behavior of the RMTR method is illustrated in \cref{fig:behavior_smiley} on the right.

We also remark that the TR method is significantly more sensitive to the choice of the initial guess than the RMTR method. 
The relative standard deviation of the obtained results varies from~$30\%$ to~$40\%$ for the TR method. 
In contrast, the relative standard deviation for the RMTR method decreases with the number of levels and it is below~$3.5\%$ for networks with~$6$ levels for both datasets. 
{The reduced sensitivity to the initial guess was also observed for multilevel parameter initialization strategy applied in the context of layer-parallel training in~\cite{cyr2019multilevel}.}

\begin{table}
  \centering
    \small
  \caption{The average total computational cost required by the deterministic TR and the RMTR method using Smiley, and Spiral datasets.
   {The results are reported in terms of fine-level work unit $W^L$.}
    Both methods employ the LSR1 scheme in order to approximate the Hessian.
    The results are obtained by averaging~$10$ independent runs. 
    The symbol~$--$ indicates that no convergence was reached within~$1,000 \ W^L$. }
  \label{table:deterministic_test}
  \begin{tabular}{ c|l|lllllll}
       \multirow{2}{*}{{{Example}}}               &         \multicolumn{1}{c|}{{  \multirow{2}{*}{{Method}}}}          &     \multicolumn{4}{c}{{Levels (Residual blocks)}}     \\         
                     &  &\multicolumn{1}{c}{{3 (25)}}  & \multicolumn{1}{c}{{4 (49)}}                 & \multicolumn{1}{c}{{5 (97)}}                  & \multicolumn{1}{c}{{6 (193)}}                                         \\ \hline \hline
    \multirow{3}{*}{{{Smiley}}} &         {TR}   & ${383.9  \pm}  {165.1}$ & $ {618.4 \pm} {271.9}$  & $ {828.4 \pm}  {397.4}$  & \textcolor{white}{0000}${--}$                  \\
& {RMTR-V} & $\textcolor{white}{0}{68.2 \pm}{5.4}$   & $\textcolor{white}{0}{82.6 \pm} {8.3}$    & ${121.2 \pm} {15.8}$ & ${133.1 \pm}{15.9}$    \\    
& {RMTR-F} & $\textcolor{white}{0}{63.4 \pm} {8.8}$   & $\textcolor{white}{0} {29.1 \pm} {1.7}$    & $\textcolor{white}{0}{19.1  \pm} {1.1} $   & $\textcolor{white}{0}{14.2  \pm} {0.4}$    \\   \hline 

    \multirow{3}{*}{{{Spiral}}} & {TR}   & $  {157.8 \pm} {52.1}$ & $ {231.3 \pm} {80.9}$   & $ {332.2  \pm}  {119.6} $ & $ {412.3  \pm}  {148.4}$ \\
					& {RMTR-V} & $  \textcolor{white}{0}{33.1 \pm} {4.1}$ & $\textcolor{white}{0}{39.9 \pm} {5.3}$ & $ \textcolor{white}{0}{83.1 \pm} {12.5}$ & ${126.8  \pm} {11.4}$ \\
					& {RMTR-F} & $  \textcolor{white}{0}{58.2 \pm} {5.2}$ & $\textcolor{white}{0}{ 28.9  \pm} {1.15}$ & $ \textcolor{white}{0}{21.7 \pm} {0.9}$ & $\textcolor{white}{0}{16.7  \pm} {0.5}$  \\  
  \end{tabular}
\end{table}

\begin{figure}[t]
\centering
\includegraphics{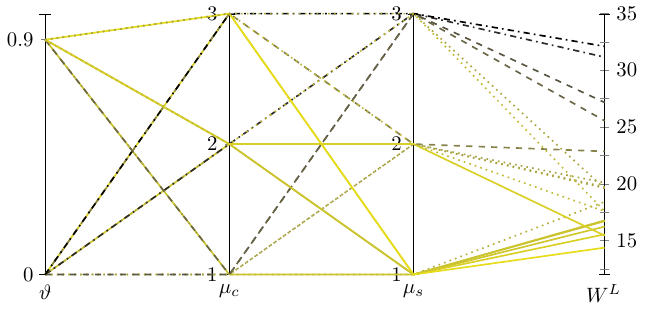}
\caption{{The computational cost of the RMTR method with respect to the choice of the momentum parameter ($\vartheta$), and varying number of the coarse-level/smoothing steps ($\mu_c/\mu_s$). 
The experiments performed using F-cycle of RMTR method with $6$ levels and Smiley dataset.}}
\label{fig:parallel_plot_rmtr_params}
\end{figure}

\paragraph{Momentum and number of coarse-level/smoothing steps}
{Next, we investigate how the number of coarse-level/smoothing steps and the use of momentum affect the computational cost of the RMTR method. 
We consider the number of coarse-level/smoothing steps from a set $\{1, 2, 3\}$. 
For the momentum parameter $\vartheta$, we investigate values $0.9$ and $0.0$, where $\vartheta=0.0$ is equivalent to not turning of the momentum term, recall \cref{sec:momentum_ml}. 
\cref{fig:parallel_plot_rmtr_params} illustrates the obtained results by means of parallel coordinate plot. 
As we can observe, the use of momentum is beneficial as it decreases the overall computational cost, i.e., the experiments, which employ the momentum parameter require a smaller number of work units (the right y-axis), which is also depicted by the use of yellow line color.
Furthermore, we notice that the increasing number of smoothing steps increases the computational cost of the RMTR method. 
This is not surprising, as the gradient evaluations on the finer levels are more expensive than on the coarser ones. 
Thus, in order to decrease the computational cost, we should delegate as many computations as possible to the coarser levels. 
Interestingly, we also see that increasing the number of coarsest-level steps is beneficial, but it does not influence heavily the computational cost of the RMTR method. 
This is due to the fact that the multilevel trust-region radius update mechanism, described in~\cref{sec:rmtr_resnet}, ensures that the size of the prolongated coarse-level correction does not exceed the finest-level trust-region radius. 
Thus, the number of the coarse-level steps is adjusted inherently by the RMTR algorithm during the training.}

\subsubsection{Hybrid (stochastic-deterministic) settings}
In this section, we compare the performance of the DSS-TR method and the DSS-RMTR method (F-cycle). 
The performed study considers three different initial mini-batch sizes~$\text{mbs}_0$, which are reset to their initial value every time a new level is taken into consideration.
More precisely, the parameter~$\text{mbs}_0$ takes on a value from~$\{ 250, 500, 1,000 \}$. 
\cref{table:stochastic_test_lsr1} reports the obtained results.
As we can see, hybrid trust-region methods perform better than their deterministic counterparts. 
We also note that the hybrid methods are computationally cheaper when the mini-batch size is initialized to a smaller value. 
This is due to the fact that small-batch methods tend to be more exploratory, which allows them to escape plateau regions.
We also highlight the fact that hybrid methods are less sensitive to the choice of the initial guess than deterministic methods.

The obtained results imply that the DSS-RMTR method performs significantly better than the DSS-TR method, in terms of total computational cost and the sensitivity to the initial guess. 
Similarly to the results obtained for the deterministic methods, the total computational cost of the DSS-TR method increases with network depth. 
In contrast, the total computational cost of the DSS-RMTR method decreases with network depth and the number of levels. 
For example, for the Spiral dataset with~$6$ levels, $\text{mbs}_0=250$ and~$193$ residual blocks, the DSS-RMTR method requires approximately~$5$ times lower computational cost than the DSS-TR method.

\begin{table}
  \centering
\small
          \caption{The average total computational cost of the DSS-TR and DSS-RMTR methods required for training dense ResNets.
     {The results are reported in terms of fine-level work unit $W^L$. }          
  Both methods employ the LSR1 scheme in order to approximate the Hessian.
The results are obtained by averaging~$10$ independent runs. }
  \label{table:stochastic_test_lsr1}
  \begin{tabular}{ c|c|l|lllcc}
        \multirow{2}{*}{{{Example}}}             &          \multirow{2}{*}{{{$\text{mbs}_0$}}}         &          \multirow{2}{*}{{{Method}}}  &     \multicolumn{3}{c}{{Levels (Residual blocks)}}     \\     
                     &            & & \multicolumn{1}{c}{{2 (13)}}                &  \multicolumn{1}{c}{{4 (49)}}                &  \multicolumn{1}{c}{{6 (193)} }                \\ \hline \hline
    \multirow{6}{*}{{{Smiley}}} & \multirow{2}{*}{{{250}}}   & {DSS-TR}          & ${20.1\pm} {1.0}$    & ${21.5\pm}  {0.9}$   & ${23.1\pm}  {1.2}$        \\
                              &                          			& {DSS-RMTR}        & ${11.7\pm}  {0.3}$    & $\textcolor{white}{0}{5.4\pm}  {0.1}$   & $\textcolor{white}{0}{4.5 \pm}  {0.1}$            \\ \cline{2-6}
                              
                              & \multirow{2}{*}{{{500}}}   	& {DSS-TR}          & ${25.2\pm} {1.6}$    & ${25.8 \pm} {1.3}$  & ${26.2\pm} {1.4}$             \\
                              &                          			& {DSS-RMTR}        & ${16.5\pm} {0.6}$    & $\textcolor{white}{0}{6.4\pm} {0.1}$    & $\textcolor{white}{0}{5.0 \pm} {0.04}$           \\   \cline{2-6}

                              & \multirow{2}{*}{{{1,000}}} & {DSS-TR}          & ${31.6\pm}  {2.6} $  & ${33.4\pm}  {2.4}  $ & ${36.2\pm} {3.3}$        \\
                              &                          						& {DSS-RMTR}        & ${18.2\pm}{0.8}$     & $\textcolor{white}{0}{7.7\pm}{0.1}$     & $\textcolor{white}{0}{6.1 \pm} {0.1}$             \\   \hline 
    \multirow{6}{*}{{{Spiral}}} & \multirow{2}{*}{{{250}}}   	& {DSS-TR}          & ${17.2\pm} {0.8}$     & ${21.4\pm} {1.1}$   & ${23.5\pm} {1.3}$              \\
                              &                          				& {DSS-RMTR}        & ${13.8 \pm} {0.5}$ & $\textcolor{white}{0}{7.2 \pm}  {0.3}$ & $\textcolor{white}{0}{4.4 \pm}  {0.07}$           \\ \cline{2-6}

                              & \multirow{2}{*}{{{500}}}   	& {DSS-TR}          & ${31.8\pm}  {3.0}$     & ${32.4\pm} {2.0}$   & ${39.3\pm} {3.7}$             \\
                              &                          			& {DSS-RMTR}        & ${25.1\pm} {2.1}$    & ${16.3 \pm} {0.4}$  & ${13.5 \pm} {0.3}$         \\   \cline{2-6}

                              & \multirow{2}{*}{{{1,000}}} 	& {DSS-TR}          & ${34.2\pm} {3.3}$     & ${43.9\pm} {6.0}$  & ${55.4\pm} {12.9}$            \\
                              &                          			& {DSS-RMTR}        & ${23.2 \pm} {1.4}$  & ${16.0\pm} {0.5}$   & ${13.7\pm} {0.3}$             \\   
  \end{tabular}
\end{table}

\section{{Hyper-parameter search for GD and Adam methods}}
\label{sec:hyper_tunning_adam_gd}
\begin{figure}[t]
\includegraphics{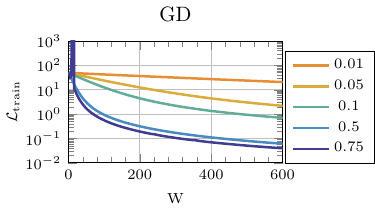}
\includegraphics{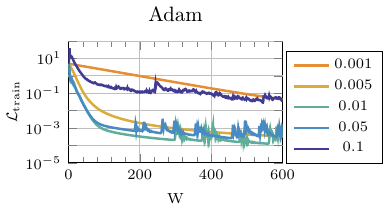}
\includegraphics{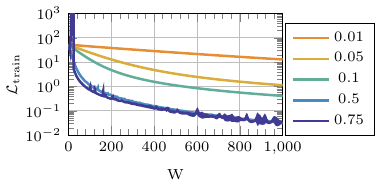}
\includegraphics{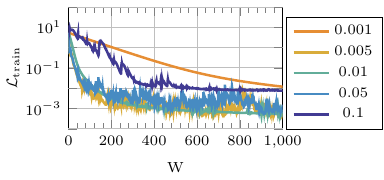}
\caption{{ Mean training loss obtained over 10 independent runs for TDD/NDR examples (\emph{Top/Bottom}) with a fixed computational buget of $600 \ W$ and $1,000 \ W$, respectively.
Experiments performed with varying learning rate using GD and ADAM methods (\emph{Left/Right}).}}
\label{fig:hyper_param_tunning_regression}
\end{figure}

{
In this section, we report the hyper-parameter selection process for GD and Adam methods employed in~\cref{sec:num_results}. 
For regression problems, we use deterministic settings.
The learning rate ($\alpha$) is sampled from the set ${ \{0.01, 0.05, 0.1, 0.5, 0.75 \} }$ and $\{0.0001, 0.005, 0.01, 0.05, 0.01 \}$ for the GD and the Adam method, respectively. 
\cref{fig:hyper_param_tunning_regression} demonstrates the obtained results for the TDD dataset on the left and NDR dataset on the right. 
As we can see, the optimal value of $\alpha$ for the GD method is $0.5$. 
The values of $\alpha$ larger than $0.5$, e.g.,~$0.75$ often cause divergence and therefore are excluded from the consideration. 
For the Adam method, we observe that the lowest value of $\pazocal{L}_{\text{train}}$ is obtained for $\alpha$ equal to $0.01$ and $0.05$, for the TDD and NDR dataset, respectively.
}

{We train the classification problems with convolutional ResNets using stochastic variants of the GD and Adam methods. 
In particular, we use mini-batch sizes of $256$ and $100$ for the Fashion and the CIFAR-10{/CIFAR-100} datasets, respectively. 
For the Fashion dataset, we sample learning rate $\alpha$ from a set $\{0.05, 0.1, 0.25, 0.5,0.75\} $ for the SGD method and from a set $\{10^{-5}, 5 \times 10^{-5}, 10^{-4}, 5 \times 10^{-4}, 10^{-3}\}$ for the Adam method.
Moreover, we consider three weight-decay strategies, denoted by $\text{WD}_{0-2}$.
More precisely, the learning rate is dropped by the factor of $0.1$ at $[50, 100, 150], [60, 120, 180]$, and $[40, 80, 120, 160]$ epochs, for $\text{WD}_0$, $\text{WD}_1$ and $\text{WD}_2$,  respectively.
For the CIFAR-10 and {CIFAR-100} datasets, we sample learning rate from a set $\{0.01, 0.05, 0.1, 0.5\} $ for SGD method and from a set ${ \{10^{-5}, 5 \times 10^{-5}, 10^{-4}, 5 \times 10^{-4} \} }$ for the Adam method.
Here, we consider four weight-decay strategies. 
For $\text{WD}_0$ and $\text{WD}_1$, we drop the learning rate by the factor of 0.1 at [50, 100, 150] and [60, 120, 180] epochs, while for $\text{WD}_2$ and $\text{WD}_3$, we drop the learning rate by the factor of 0.5 at [50, 100, 150] and [60, 120, 180] epochs. }

\begin{figure}[t]
\includegraphics{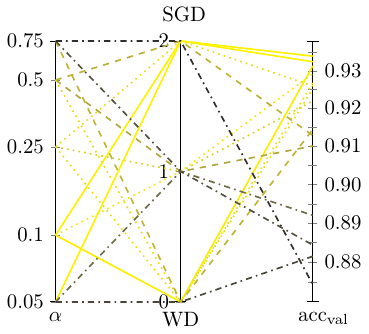}
\includegraphics{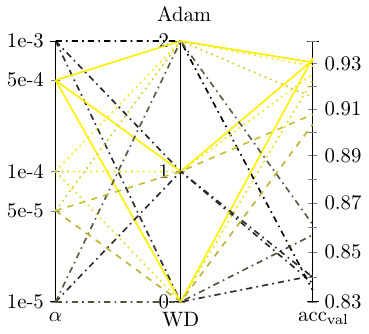}
\includegraphics{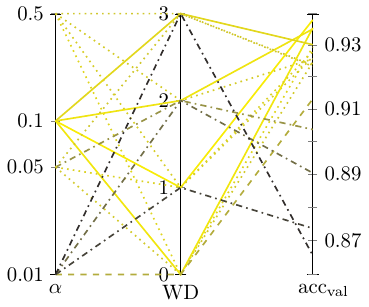}
\includegraphics{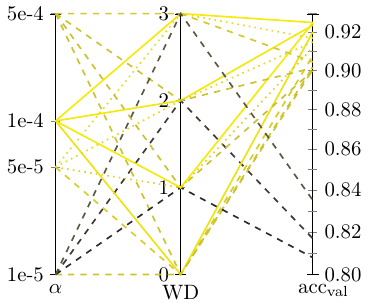}
\includegraphics{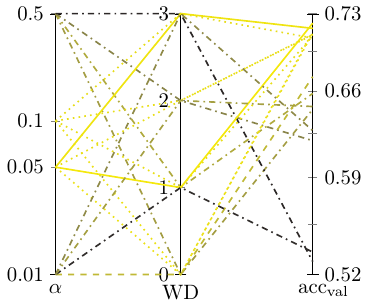}
\includegraphics{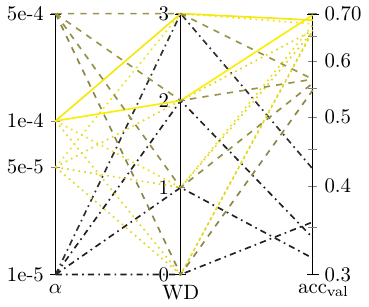}
\caption{{
The performance of SGD and Adam method (\emph{Left/Right}) in terms of the validation accuracy with respect to the choice of the learning rate ($\alpha$) and the weight-decay strategy (WD). 
The experiments performed for Fashion/CIFAR-10/{CIFAR-100} datasets (\emph{Top/{Middle/Bottom} }) and ResNets associated with level $l=2$.}}
\label{fig:hyperparam_tunning_conv}
\end{figure}

{\cref{fig:hyperparam_tunning_conv} demonstrates the obtained results in terms of parallel coordinate plots. 
Our main interest is to select hyper-parameters that yield the highest validation accuracy (acc$_{\text{val}}$). 
As we can observe, for the Fashion dataset, the SGD with $\alpha = 0.1$ and $\text{WD}_2$ strategy and Adam with $\alpha = 5 \times 10^{-4}$ and $\text{WD}_0$ strategy yield the best results.
For the CIFAR-10 dataset, the highest acc$_{\text{val}}$ is reached using SGD with $\alpha = 0.1$ and $\text{WD}_1$ strategy and Adam with $\alpha = 10^{-4}$ and $\text{WD}_3$ strategy.
{In case of CIFAR-100 dataset, the SGD method configured with $\alpha=0.05$ and $\text{WD}_1$ strategy and Adam setup with $\alpha = 5 \times 10^{-4}$ and $\text{WD}_2$ strategy provide the best results.}
Please note, that for {all three} datasets, the SGD method is able to achieve higher validation accuracy than the Adam method. 
{The largest difference can be observed for the CIFAR-100 dataset, for which the SGD method is able to achieve  acc$_{\text{val}}=72.1\%$. 
In comparison, the highest validation accuracy obtained by the Adam method equals $69.7\%$.}
}

\end{sloppypar}

\bibliographystyle{siamplain}
\bibliography{biblio}
\end{document}